\documentclass[letterpaper,journal]{IEEEtran}
\usepackage{amsmath,amsfonts}
\usepackage{algorithmic}
\usepackage{algorithm}
\usepackage{array}
\usepackage[caption=false,font=normalsize,labelfont=sf,textfont=sf]{subfig}
\usepackage{textcomp}
\usepackage{stfloats}
\usepackage{url}
\usepackage{verbatim}
\usepackage{graphicx}
\usepackage{cite}

\usepackage{graphicx}
\usepackage[table]{xcolor}
\usepackage{multirow}
\usepackage{tabu}
\usepackage{bbm}
\usepackage{diagbox}
\usepackage[english]{babel}
\usepackage{comment}
\usepackage{array}
\usepackage{amstext}
\usepackage{makecell}
\usepackage{caption}
\usepackage{tablefootnote}
\usepackage{babel}
\usepackage{booktabs} 
\usepackage{color}
\usepackage{amsmath}
\usepackage{wrapfig}
\usepackage{amssymb}
\usepackage{hyperref}
\usepackage{nameref}
\usepackage{marvosym}

\newif\ifdraft
\draftfalse
\drafttrue

\definecolor{burntorange}{rgb}{0.8, 0.33, 0.0}
\definecolor{orange}{rgb}{1,0.5,0}
\definecolor{green0}{rgb}{0.1,0.7,0.1}

\ifdraft
\newcommand{\PF}[1]{{\color{red}{\bf PF: #1}}}

\newcommand{\JC}[1]{{\color{blue}{\bf JC: #1}}}

\newcommand{\HT}[1]{{\color{blue}{\bf HT: #1}}}

\newcommand{\AF}[1]{{\color{green0}{\bf AF: #1}}}

\newcommand{\NE}[1]{{\color{violet}{\bf NE: #1}}}

\newcommand{\RL}[1]{{\color{orange}{\bf RL: #1}}}

\newcommand{\HL}[1]{{\color{burntorange}{\bf HL: #1}}}

\newcommand{\JX}[1]{{\color{brown}{\bf JX: #1}}}

\newcommand{\TODO}[1]{\textbf{\color{red}[TODO: #1]}}
\else
\newcommand{\PF}[1]{}

\newcommand{\JC}[1]{}

\newcommand{\HT}[1]{}

\newcommand{\AF}[1]{{\color{green0}{}}}

\newcommand{\NE}[1]{{\color{violet}{}}}

\newcommand{\RL}[1]{{\color{orange}{}}}

\newcommand{\HL}[1]{{\color{pink}{}}}

\newcommand{\JX}[1]{{\color{brown}{}}}

 \newcommand{\TODO}[1]{}
 
\fi

\newcommand{\parag}[1]{\noindent{\textbf{#1}}}
\newcommand{\sparag}[1]{\textbf{#1}}

\newcommand{\acron}{\emph{GenMed}}
\newcommand{\acronN}{\emph{GenMed-Naive}}
\newcommand{\acronF}{\emph{GenMed-Full}}
\newcommand{\acronB}{\emph{GenMed-Base}}

\begin{document}

\title{GenMed: A Pairwise Generative Reformulation of Medical Diagnostic Tasks}

\author{
Hantao Zhang$^*$,
Weidong Guo$^*$,
Yuhe Liu$^*$,
Jiancheng Yang,
Sathvik Bhagavan,
Danli Shi,
Mingda Xu,\\
Pascal Fua\textsuperscript{\Letter},~\IEEEmembership{Fellow,~IEEE}
\thanks{
Hantao Zhang, Sathvik Bhagavan, Mingda Xu, and Pascal Fua are with the CVLab, École Polytechnique Fédérale de Lausanne (EPFL), Switzerland. Weidong Guo is with Fudan University, Shanghai, China. Yuhe Liu is with Beihang University, Beijing, China. Jiancheng Yang is with ELLIS Institute Finland, Finland, and also with Aalto University, Finland. Danli Shi is with The Hong Kong Polytechnic University, Hong Kong SAR, China. $^*$Equal Contribution. \textsuperscript{\Letter}Corresponding Author: Pascal Fua (E-mail: pascal.fua@epfl.ch).
}
}

\maketitle
\newcommand{\revise}[1]{\textcolor{red}{#1}}

\newcommand{\xd}[0]{{{{\cal X}_{\Theta} }}}
\newcommand{\yd}[0]{{{{\cal Y}_{\Theta} }}}
\newcommand{\fx}[0]{{f_x}}
\newcommand{\fy}[0]{{f_y}}
\newcommand{\loss}[0]{{{{\cal L}_{\Theta} }}}

\newcommand{\dstep}[0]{{{{\cal S}_{\Theta} }}}
\newcommand{\dfull}[0]{{{{\cal D}_{\Theta} }}}
\newcommand{\dfulx}[0]{{{{\cal D}^x_{\Theta} }}}


\begin{abstract}

Data-driven medical AI is traditionally formulated as a discriminative mapping from input $X$ to output $Y$ via a learned function $f$, which does not generalize well across heterogeneous data and modalities encountered in real-world clinical settings. In this work, we propose a fundamentally different, generative paradigm. We model the joint distribution $P(X,Y)$ using diffusion models and reframe inference as a test-time output optimization problem. By guiding the generative process to match observed inputs, our framework enables flexible, gradient-based conditioning at inference time without architectural changes or retraining, effectively supporting arbitrary and previously unseen combinations of observations. Extensive experiments demonstrate strong performance across standard and cross-modality medical image segmentation, few-shot segmentation with only 2 or 4 training samples, degraded-input segmentation, shape completion from sparse and partial observations, and zero-shot application to demonstrate generality. To support these evaluations, we curated and released a large-scale text–shape dataset derived from MedShapeNet. Our results highlight the versatility of generative joint modeling as a foundation for reusable, task-agnostic medical AI systems.

\end{abstract}



\section{Introduction}

Data-driven AI now permeates nearly all aspects of healthcare~\cite{topol2019high,moor2023foundation}, spanning the entire clinical pipeline—from diagnosis~\cite{jeong2025reducing} to treatment~\cite{wang2025clinical} and prognosis~\cite{alum2025ai}. Within the diagnostic workflow, a crucial step is the delineation of organ contours and their refinement into accurate anatomical shapes, which is essential for downstream diagnosis and surgical planning~\cite{song2012optimal}. In the actual clinical workflow, this process typically implicitly involves both segmentation and shape completion tasks~\cite{zhu2025improving,zhang2025tuning,li2025medshapenet}. Segmentation provides an initial estimate of organ or tumor regions~\cite{zhang2025tuning,liu2023clip}, while shape completion refines the results by correcting over- and under-segmentation errors to produce accurate shape representations. A classic example is the segmentation and reconstruction of small anatomical structures in medicine, such as the reconstruction of pulmonary segments ~\cite{xie2025template}. Most current AI approaches to implementing clinical diagnostic pipelines rely on modeling conditional probabilities $P(Y|X)$ of an output $Y$ given an input $X$. For example, in medical image segmentation, $X$ denotes an input image and $Y$ the corresponding segmentation mask. Similarly, when performing shape completion, $X$ denotes the partial shape and $Y$ the completed one. $P(Y|X)$ is usually learned from training data and then used at inference time. This is true of both {\it discriminative} methods using traditional feed-forward networks such as  nnU-Net~\cite{isensee2021nnu} and more recent {\it generative} ones that rely on conditional diffusion models, such as~\cite{wu2024medsegdiff}. 

While this is highly effective in narrow, well-controlled settings, this places a heavy burden on the training process. In complex real-world situations, models must handle heterogeneous data across centers, modalities, and tasks, as exemplified by the {\it multi-X} setting of~\cite{yang2024multi}. Recent prompt-based approaches~\cite{kirillov2023segment,ma2024segment} improve flexibility by conditioning on structured prompts or auxiliary inputs. However, this requires large-scale datasets and extensive augmentation. As a result,  proper generalization to unseen prompt formats or novel observation modalities is never guaranteed at inference time.

In this work, we advocate a fundamentally different strategy. Instead of explicitly learning the conditional distribution $P(Y|X)$,  we use a generative diffusion framework~\cite{ho2020denoising,rombach2022high} to model the joint distribution $P(X,Y)$. At inference time, given an input $X^*$ we then guide the generative process toward a plausible pair $(X', Y')$ such that $X' \approx X^*$. In our experiments, we show that this approach delivers reliable results even when $X^*$ lies outside the training distribution of $X$, such as when using a model trained on CT data to process MRI data without re-training. We will refer to this generative method as \acron{}. Because training only involves the unconditional modeling of the distributions of $(X, Y)$ pairs, while inference relies on gradient-based guidance for search, \acron{} eliminates the need for a dedicated input-conditioning encoder and naturally supports arbitrary combinations of observations at test time. In effect, we can train once and then test with any conditioning signal without fine-tuning. 

While jointly modeling multiple variables is common in diffusion models,  existing approaches to doing it take a tack that is different from ours. Some model both images and their corresponding masks to learn better representations in an unsupervised manner~\cite{sauvalle2024hybrid}, while others adopt a multi-task learning perspective to model joint distributions~\cite{Li25d}. However, none of them explicitly reformulates the relationship between input and output variables $(X, Y)$ within a single task. More importantly, they remain within the input-conditional diffusion paradigm, which implicitly assumes that $X^*$ lies within the training distribution of $X$. In medical settings, however, due to device failures, signal interference, and other unpredictable sources of error, $X^*$ is often out-of-distribution relative to the training data $X$ and we will see that our method has a distinct advantage in that scenario. Moreover, modeling $P(X,Y)$ rather than $P(Y \mid X)$ makes it possible to leverage richer information during training under limited data conditions, which aligns well with practical clinical needs.

We introduced an early version of \acron{} in a conference paper~\cite{Zhang25b}, but only for segmentation and using a comparatively weak approach to guidance. In this extended journal version, we propose a much improved approach to guidance and broaden the scope of our method to include tasks that, unlike segmentation, require a latent vector representation. We chose to focus on these two tasks because they jointly constitute the backbone of numerous diagnostic applications, such as assessing {\it bronchiectasis}~\cite{polverino2017european}, that is, structural anomalies such as airway narrowing and dilation, or determining tumor invasion depth in rectal cancer staging~\cite{zhang2025tuning}. While segmentation is inherently prone to errors due to the limited resolution of medical imaging, these tasks require an accurate structural representation. Thus, shape completion serves as an essential refinement step that turns raw segmentation masks into consistent anatomical shapes~\cite{li2025medshapenet}.

\begin{itemize}

	\item We apply a model trained on 3D volume segmentation to low-resolution volumes and even single 2D slices at test time, without retraining or specialized augmentation, effectively leveraging the intrinsic image priors learned from the original data.
	
	\item We demonstrate that a single model trained only on complete shapes can perform shape completion from diverse observations, including single-plane, multi-plane, tri-plane, and noisy or partial inputs.

        \item We combine our segmentation and shape-completion algorithms into a unified pipeline that generates fine-grained shapes directly from raw medical images, demonstrating the benefit of leveraging the \acron{} paradigm to support the medical diagnosis workflow.

\end{itemize}

\begin{figure*}[tb]
    \centering
    \includegraphics[width=1\linewidth]{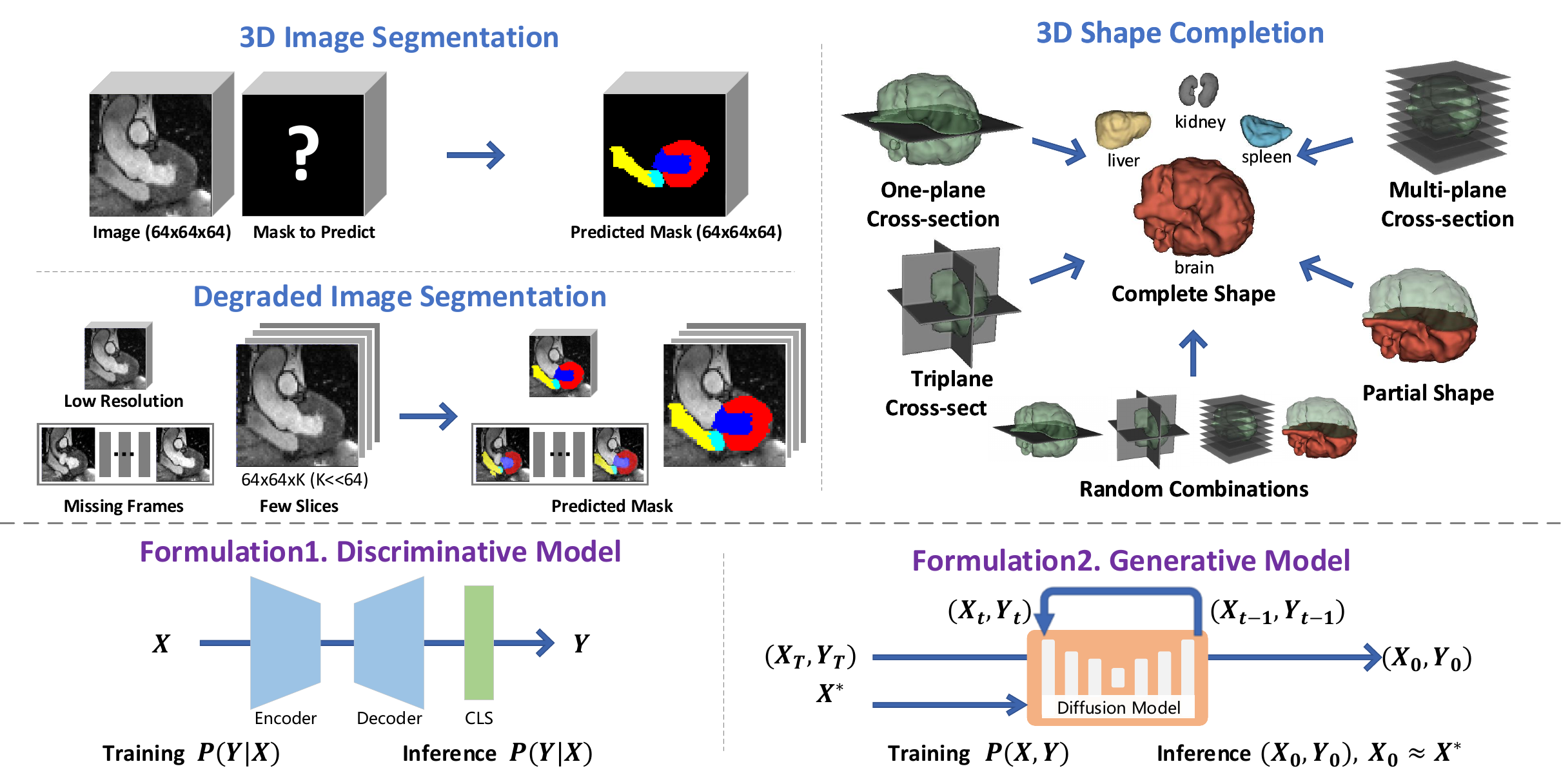}
    \caption{\textbf{A single generative model for diverse medical tasks.} We address tasks closely related to medical diagnosis: (1) \textbf{3D image segmentation}, including standard segmentation on complete 3D medical images, as well as clinically common yet challenging scenarios such as zero-shot cross-modality segmentation, few-shot segmentation with only 2 or 4 training samples, and degraded-input segmentation on images with downsampling or missing slices; (2) \textbf{Shape completion} from partial voxel cues (e.g., single-, multi-, or orthogonal-plane cross-sections). Unlike the discriminative paradigm, which only estimates the conditional distribution $P(Y \mid X)$, our generative formulation models the joint distribution $P(X, Y)$ and performs constraint-guided sampling to enforce input conditions during inference.}
    \vspace{-10px}
    \label{fig:compare}
\end{figure*}

\IEEEpubidadjcol


\section{Related Works}

Diagnostic applications often rely on segmentation algorithms to extract the rough 3D shape of organs, followed by shape completion to remove artefacts and precisely model geometric relationships between organs~\cite{li2025medshapenet}. \acron{} is intended to perform both these tasks using the same generative framework, which involves  the modeling of the joint distribution $P(X, Y)$ instead of the conditional distribution $P(Y \mid X)$, where $X$ represents an input and $Y$ the desired output. In this Section, we briefly review current approaches to medical image segmentation and shape completion. We also discuss existing approaches to generative modeling of joint distributions and to guiding such approaches.

\subsection{Medical Image Segmentation}

Medical Segmentation often involves discriminative prediction of the ground truth label for each pixel using forward model architectures such as U-Net~\cite{ronneberger2015u} and nnU-Net~\cite{isensee2021nnu}. In recent years, many different methods have been proposed to better capture fine-grained details and improve robustness to domain shifts. These include attention mechanisms~\cite{wang2022uctransnet}, multi-scale feature extraction~\cite{kushnure2021ms}, and adversarial training~\cite{nie2020adversarial}. Despite these efforts, substantial challenges remain, particularly for complex anatomical structures with only limited annotations~\cite{feng2021interactive}, multi-modal cross-domain settings~\cite{ding2022cross}, and corrupted or degraded imaging conditions~\cite{kucs2024medsegbench}.

Some approaches, such as~\cite{wu2024medsegdiff, chen2024hidiff, zhang2025diffuseg} add generation in the form of masks generated by image-guided diffusion. However, the underlying paradigm remains fundamentally discriminative. Such methods depend heavily on the input format for feature extraction and decision-making, which in turn requires large-scale datasets for effective generalization. To help with this some large-scale datasets such as TotalSegmentator (TS) dataset~\cite{zhuang2019evaluation} and MedTrinity~\cite{xiemedtrinity} have been built. However, this approach faces practical limitations: a large number of anatomical structures to annotate, the diversity of imaging modalities, and privacy constraints make it difficult to achieve truly comprehensive data coverage, which in turn hampers the training of highly robust models. To reduce annotation cost, labeling strategies, such as SAROS~\cite{koitka2024saros}, adopt sparse labeling. Another approach is to focus on constructing specialized, domain-specific datasets for a particular body region or modality, e.g., abdominal datasets~\cite{li2024abdomenatlas} or rectal cancer datasets~\cite{zhang2025tuning}. However, more often than not, such datasets are unavailable, and segmentation performance degrades in  complex cross-domain input scenarios~\cite{ azad2024medical}. Using atlases~\cite{chen2010image, bai2015bi, ding2022cross} mitigates these difficulties but faces limitations of its own, including  limited scalability, difficulty in capturing fine-grained structures, and reliance on labor-intensive, domain-specific atlas construction. We will demonstrate that \acron{} alleviates all these problems. 

\subsection{Shape Completion}
3D shape generation is typically built upon various representations, such as dense voxelized SDFs on regular grids~\cite{cheng2023sdfusion,li2023diffusion} and octree-based adaptive latent representations~\cite{xiong2025octfusion}. Regardless of the underlying representation, recent shape-completion approaches share a common paradigm: the incomplete shape is treated as a prompt and incorporated into the training process of the diffusion model~\cite{zhang2026high,wang20253d}. However, as in the segmentation case, this relies heavily on the specific format of the prompt~\cite{cheng2023sdfusion} and requires large-scale datasets for effective learning~\cite{chen2025dora, zhao2025hunyuan3d}. In medical settings, obtaining the necessary large-scale, well-curated, and clean datasets is particularly difficult~\cite{sun2022topology, xie2025template}. Although recent efforts such as MedShapeNet~\cite{li2025medshapenet} have produced large-scale shape datasets for the medical domain, the inherent difficulty of collecting and curating medical data makes it challenging to achieve per-category sample sizes comparable to natural image datasets such as ShapeNet~\cite{chang2015shapenet}—a gap that limits the generalizability of trained models to new, unseen cases. Furthermore, due to the unique characteristics of certain anatomical structures, including fine-grained geometry and high structural complexity~\cite{gafencu2024shape,xie2025efficient}, the corresponding datasets~\cite{li2025medshapenet} often contain outlier noise points, corrupted shape artifacts, and incomplete annotations. These data quality issues further increase training difficulty and hinder model adaptation to new cases. Therefore, effective adaptation strategies for shape completion on novel cases remain under-explored and warrant further investigation. 

\subsection{Generative Modeling}
\label{sec:generative}
Generative models based on diffusion~\cite{Song21c} and flow-matching~\cite{Lipman22} schemes have become popular for such tasks, and a range of \emph{classifier-free guidance} (CFG) variants have been proposed to improve their controllability. \textit{Schedule-based} variants replace the constant guidance weight with a time-dependent schedule; Wang \emph{et al.}~\cite{xi2024analysis} show that monotonically increasing schedules (Linear, power-cosine, exponential) outperform the constant baseline by suppressing excessive guidance during early denoising steps. \textit{Update-rule} variants instead modify the CFG direction itself: APG~\cite{sadat2024eliminating} down-weights the parallel component of the update to mitigate oversaturation; Rectified-CFG++~\cite{sainirectified} uses a predictor--corrector integrator to keep flow trajectories on the conditional transport path; and CFG-Zero$^\star$~\cite{fan2025cfg} corrects under-fitted velocity estimates via an optimized scale and a zero-init first ODE step. All are training-free, drop-in replacements for vanilla CFG.

A parallel line of work focuses on guiding the model's path at inference time. Early approaches~\cite{dhariwal2021diffusion,ho2022classifier} modulate the reverse diffusion process to enable conditional generation. Subsequent works extended this idea to finer-grained control across diverse applications, including 3D shape design~\cite{you2025physgen}, autonomous driving~\cite{zhengdiffusion,yang2024diffusion}, and molecular generation~\cite{choi2024pidiff}. While these methods differ in their specific strategies -- for instance, D-Flow~\cite{ben2024d} optimizes the initial noise over the entire generation trajectory using a differentiable ODE solver, whereas others~\cite{song2023loss,yang2024diffusion} treat guidance as a per-step search problem on intermediate states and exploit the gradient of a guidance loss to steer the denoising process -- they share a common principle: leveraging gradient-based optimization, tailored and adapted to the specific task at hand. Despite these differences, all such approaches steer \emph{output} generation given a fixed \emph{input} conditioning signal, leaving the model's input--output contract unchanged. Our work moves away from such rigid input-conditioned discriminative modeling. By adopting a pair-based paradigm and reformulating the problem as a generative task, our method operates entirely at test time by steering the search process of a diffusion model~\cite{singhalgeneral}, without updating any model parameters. In contrast to prior diffusion‑based approaches that model the joint distributions of input‑output pairs~\cite{sauvalle2024hybrid}, we do not merely leverage such pairs for representation learning to obtain better feature extraction. Instead, we steer the model’s behavior at test time. We will show that this increases robustness and adaptability.


\section{Method}

\newcommand{\dstex}[0]{{{\cal S}^x_{\Theta} }}
\newcommand{\encd}{{\mathrm{Enc}}}
\newcommand{\decd}{{\mathrm{Dec}}}
\newcommand{\decdx}[0]{{\mathrm{Dec}^x}}

\newcommand{\tX}[0]{{\tilde{X}}}
\newcommand{\tY}[0]{{\tilde{Y}}}


\begin{figure*}[tb]
        \centering
	\includegraphics[width=1\linewidth]{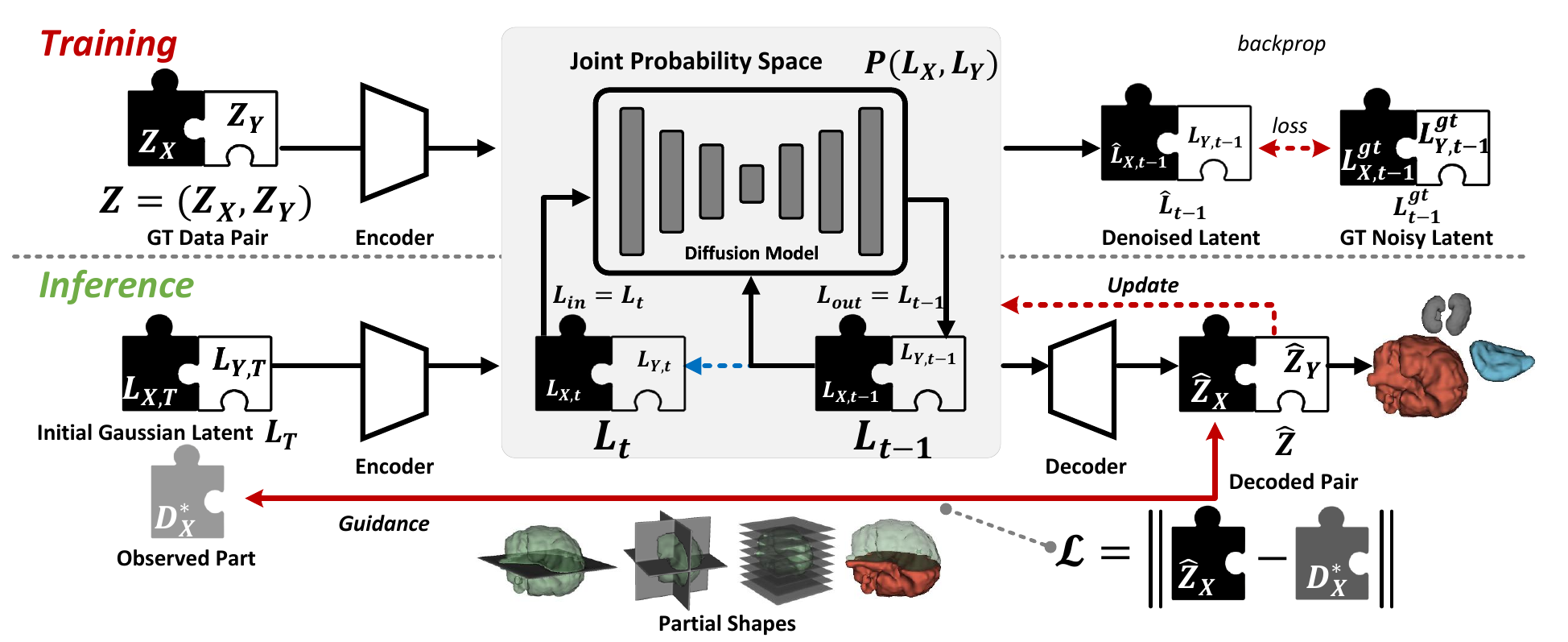}
    \caption{\textbf{\acron{} Architecture.} During training, \acron{} learns to denoise samples from the joint distribution $ P(X, Y) $ in either explicit or latent space. At inference, the known condition $ X^* $ is used to explicitly or implicitly guide the sampling process, ensuring that the generated outputs $X_0$ adhere to the constraint $ X_0 \approx X^* $.}
    \vspace{-10px}
	\label{fig:model}
\end{figure*}

\acron{} is predicated on treating both the traditional \emph{input} and \emph{output} of a discriminative model as the joint variables modeled by a single generative model.  This enables a fundamental form of generalization, which we will demonstrate in the experiment section. At  test time, steering can be used not merely to refine outputs, but to \emph{repurpose} the model for distinct tasks within a unified framework. We first present our approach when $X$ and $Y$ are variables of immediate interest. For example, $X$ can be an image and $Y$ a segmentation mask. We then extend it to the case where $X$ and $Y$ are entangled components of a latent vector that, taken together, can be decoded into the variable of direct interest $Z$, such as a 3D shape. These two scenarios amount to operating either in an explicit space or in a latent space, both key paradigms in medical image analysis.

\subsection{Replacing Conditional by Pairwise Diffusion}
\label{sec:pairwise}

Formally, given a task where the input is $X$, our goal is to predict the corresponding output $Y$, which may represent a segmentation mask.  Instead of modeling the conditional distribution $P(Y|X)$, we model the joint one $P(X, Y)$ by training a diffusion model to generate consistent $(X,Y)$ pairs. At inference time, we use the known value of $X^*$ to guide the diffusion process to produce a pair $(X',Y')$ such that $X' \approx X^*$ and $Y'$ is therefore consistent with $X$. We now formalize the diffusion scheme we use to model the joint distribution $P(X,Y)$ and introduce two separate ways to provide the guidance.

\subsubsection{Unguided Diffusion}
\label{self:unguided}
To generate a clean pair $(X_0, Y_0)$ from Gaussian random variables 
$(X_T, Y_T) \sim \mathcal{N}(0, I)$, we follow the standard DDPM 
framework~\cite{ho2020denoising}. A noise-prediction network is trained on the 
forward process
\begin{align}
X_t = \sqrt{\bar{\alpha}_t}\, X_0 + \sqrt{1 - \bar{\alpha}_t}\, \epsilon, 
\quad \epsilon \sim \mathcal{N}(0, I),
\end{align}
where $\bar{\alpha}_t = \prod_{s=1}^{t} \alpha_s$ follows a predefined 
noise schedule. Based on this trained network, we define a reverse-step 
operator $\dstep$ that maps a noisier state to a less noisy one:
\begin{align}
 (X_{t-1}  , Y_{t-1}) &= \dstep (X_t,Y_t) \; , \label{eq:step} 
\end{align}
for $t$ in $[T,1]$. Given a decreasing sequence of time steps 
$T = t_N > t_{N-1} > \cdots > t_1 > t_0 = 0$, we iteratively apply 
Eq.~\ref{eq:step} from $t = N$ down to $t = 1$. We denote the result of 
the full denoising sequence as
\begin{align}
(X_0, Y_0) &= \dfull(X_T, Y_T) \; .
\label{eq:diffusion}
\end{align}

In the remainder of the paper, we will take $\dfulx$ to be the part of $\dfull$ that returns $X_0$ only, so that we can write $X_0 = \dfulx(X_T,Y_T)$. We define $\dstex$ similarly. 

\subsubsection{Naive Guidance}
\label{sec:naive}

In our original conference paper~\cite{Zhang25b}, we proposed providing guidance by replacing $X_t$ in Eq.~\ref{eq:step} by a noisy version $X_t^*$ of the target $X^*$, where $X_t^* = \sqrt{\bar{\alpha}_t}\, X^* + \sqrt{1-\bar{\alpha}_t}\,\epsilon$. The update equations we perform at each step of the denoising process become
\begin{align}
(X_{t-1}, Y_{t-1}) &= \dstep(X_t^*, Y_t). \label{eq:hardguiduance}
\end{align}
This works as $X_t^*$ provides a useful search direction toward the final target, but we found experimentally that simply replacing $X_t$ by $X_t^*$ distorts the diffusion process too much and sometimes results in suboptimal performance.

\subsubsection{Improved Guidance}
\label{sec:improved}

To steer the diffusion process towards producing an $X_{0}$ output that is close to the desired $X^*$ without undue distortions, we designed a {\it softer} approach to guidance.  We introduce a loss function
\begin{equation}
\loss_(X_t, Y_t) =  \| \dstex(X_{t}, Y_{t}) - X^*_{t-1} \|^2 \; , \label{eq:loss1}
\end{equation}
We then  rewrite the updates of  Eq.~\ref{eq:hardguiduance}  as
\begin{gather}
 \tX_{t} = X_t - \eta \, \nabla_{X_t} \loss(X_t, Y_t) \; ,\nonumber \\
 \tY_{t} = Y_t - \eta \, \nabla_{Y_t} \loss(X_t, Y_t) \; , \label{eq:softguidance} \\
 (X_{t-1}, Y_{t-1}) = \dstep(\tX_t, \tY_t) \; . \nonumber
\end{gather}
Unlike naive guidance, which applies explicit guidance solely to $X_{t}$ while neglecting $Y_{t}$ and thereby introduces a semantic gap between the paired outputs, our approach updates both $X_{t}$ and $Y_{t}$ simultaneously, mitigating this inconsistency. Since $X$ and $Y$ are processed as separate channels by the network that implements the $\dstep$ denoising function of Eq.~\ref{eq:step}, even though we only supervise on $X_{t-1}$, it still back propagates through the shared representation and updates the parts
of the model that affect $Y_t$. An interesting observation is that, when operating in the explicit space, the proposed improved guidance is fully compatible with the naive guidance discussed in Sec.~\ref{sec:naive}. In this case, the update of $X_t$ in Eq.~\ref{eq:softguidance} can be simplified by directly replacing it with $X_t^*$ from Eq.~\ref{eq:hardguiduance}. Therefore, the complete update rule can be written as follows:
\begin{gather}
\loss(X_{t}, Y_{t}) = \left\| \dstex(X_t^*, Y_t) - X^*_{t-1}
\right\|^2, \nonumber \\
\tilde{X}_t = X_t^*, \quad
\tilde{Y}_t = Y_t - \eta \nabla_{Y_t} \loss(X_t, Y_t), \label{eq:softguidance_explicit} \\
(X_{t-1}, Y_{t-1}) = \dstep(\tilde{X}_t, \tilde{Y}_t). \nonumber
\end{gather}

\subsection{Pairwise Diffusion for Latent Models}
\label{sec:latent}

When working with a latent vector model, the diffusion process operates on the encoded latent vectors rather than on the real-world objects directly. Let $Z$ denote a real-world object, such as a 3D mesh. An encoder $\encd$ maps $Z$ to a latent vector $L$, and a decoder $\decd$ maps the latent vector back to a reconstructed object $D$:
\begin{align}
L & = \encd(Z) \; , \label{eq:latent} \\
D & = \decd(L) \; , \nonumber
\end{align}
with $D \approx Z$ once $\encd$ and $\decd$ are properly trained. We write $\decdx$ for the part of $\decd$ that returns only the $X$ component.

To frame this in the pairwise setting, $Z$ (e.g., a 3D mesh) can be partitioned into a pair $(Z_X, Z_Y)$ in the real-world (explicit) space, where $Z_X$ is the observed shape fragment of the damaged part at the time of repair, and $Z_Y$ is the portion to be completed.We denote by $L_X$ and $L_Y$ the latent counterparts of $Z_X$ and $Z_Y$, which take the role of the vectors $X$ and $Y$ from Section~\ref{sec:pairwise}. When we wish to add guidance as in Section~\ref{sec:improved}, two difficulties arise:
\begin{enumerate}
    \item While the $X$ and $Y$ components can be partitioned in the explicit space, it is in general, extremely difficult to do this in latent space, i.e., partition the latent vector to match both components. As a result, $L_X$ and $L_Y$ cannot be straightforwardly localized or manipulated separately, and neither can be decoded on its own.
    \item The decoder is trained only on clean latents and has never seen noisy ones. Hence the naive formulation
    \begin{equation}
        D_{t} = \decd(L_{t}) \; ,
    \end{equation}
    which attempts to decode an intermediate noisy latent, falls outside the decoder's training distribution. In practice, we found this formulation to be unstable and to yield poor results.
\end{enumerate}
 
We must therefore defer decoding until the end of the denoising trajectory and apply guidance only on the final reconstruction. Concretely, we let the denoising process run to completion and define the loss directly on the final decoded output:
\begin{equation}
\label{eq:latent_loss}
\loss(L_{X,t}, L_{Y,t}) = \| \decdx(\dfull(L_{X,t}, L_{Y,t})) - D_X^* \|^2 \; ,
\end{equation}
where $D^* = Z^*$ denotes the full target object and $D_X^*$ its restriction to the $X$ part. The update steps are then performed as in Eq.~\ref{eq:softguidance}.


\section{Experiments}

We now validate the two versions of our approach presented in the method section: 
\begin{itemize}

 \item the one that operates in the explicit space where the input and output variables $X$ and $Y$ are of immediate interest as described in Section~\ref{sec:pairwise},
 
 \item the other operates in the latent space with latent variables that must be decoded to represent real physical objects, as described in Section~\ref{sec:latent}.

\end{itemize}
We use medical image segmentation to test the first and shape-completion to test the second. Because these two tasks are complementary, we then integrate their implementation into a unified pipeline that generates fine-grained geometrically correct shapes directly from raw medical images, thus demonstrating the ability of \acron{}  to support a complete medical diagnosis workflow.

\subsection{Pairwise Diffusion for Image Segmentation}
\label{sec:exp_segmentation}

Since segmentation requires fine-grained, pixel-level precision to recover intricate details, it naturally relies on explicit spatial representations, which makes it a good proving ground for \acron{}, and more specifically for our original guidance approach of Section~\ref{sec:naive} and for the more sophisticated one of Section~\ref{sec:improved}. We will refer to them as  \acronN{} and  \acronF{}, respectively. Here,  the target $X$ is an image volume while  $Y$ denotes the corresponding segmentation mask. To test for robustness to imaging artifacts,  $X$ can be either the original or a degraded version of it.

\parag{Baselines.}
We compare our method against three classes of baselines. For each one, we select the most representative methods. 

\begin{itemize}

\item \textit{Discriminative Methods.} 
We chose the widely used convolutional nnU-Net~\cite{isensee2021nnu} and the transformer-based SwinUNETR~\cite{hatamizadeh2021swin}, both of which still deliver state-of-the-art performance~\cite{gunawardhana2026comprehensive,zhuang2025mim}.

\item \textit{Registration Atlas.} 
Atlas-based segmentation algorithms have evolved from early traditional approaches~\cite{bai2015bi} to more recent deep learning-based methods~\cite{vakalopoulou2018atlasnet,huang2021medical}. We chose the newest publicly available atlas-based method, CMMAS~\cite{ding2022cross}. Its search paradigm is conceptually related to ours, which makes the comparison natural. 

\item \textit{Input-Cond Diffusion.}
We use MedSegDiffv2~\cite{wu2024medsegdiff}, a classical input conditioning diffusion-based segmentation approach. Although not the latest, it remains one of the most widely used and well-evaluated diffusion-based segmentation baselines in medical image segmentation~\cite{zhang2025diffatlas,khan2025comprehensive}, validated across 20 tasks spanning diverse imaging modalities.
\end{itemize}
For a fair comparison, we augment the data in the same way to train all models. In contrast, we do not compare to approaches that focus on large-scale pre-training~\cite{jimenez2025dino} or semi-supervised learning~\cite{chen2025multi} because they introduce additional factors beyond the scope of this paper. To quantify our results, we adopt the evaluation protocol of Metrics Reloaded~\cite{maier2024metrics} and related works~\cite{ding2022cross,wu2024medsegdiff}, employing the Dice similarity coefficient (Dice) and normalized surface distance (NSD)~\cite{deepmind2018surface} as our primary metrics. 

\parag{Datasets.}
We use two datasets MM-WHS~\cite{zhuang2019evaluation} and TotalSegmentator (TS)~\cite{zhuang2019evaluation}. In both cases, we retained five labels for full heart segmentation: Myocardium of the LV (\textit{Myo}), LV blood cavity (LV), left atrium (LA), right atrium (RA), and right ventricle (RV). For the TS dataset, we used the cardiac CT portion, which contains 746 cases. For MM-WHS, we used the 20 available CTs and 20 MRIs.

\parag{Comparative Results.}
We compare \acronN{} and  \acronF{} against our baselines in different experimental settings that correspond to frequently occurring  scenarios in medical practice. In all tables below, we group the baselines  as (a) discriminative methods, (b) atlas-based method, (c) input-conditioned generative methods, and (d) the two versions of our method. The bold face numbers denote the best performing method.


\begin{table*}[htbp]

\caption{\textbf{Segmentation under Standard Settings.} }
\label{tab:seg_normal}
\centering
{\scriptsize
\begin{tabular*}{\hsize}{@{}@{\extracolsep{\fill}}clcccccccccccc@{}}
\toprule
\midrule
\multirow{2}{*}{} & \multirow{2}{*}{Methods}
& \multicolumn{2}{c}{Myo}
& \multicolumn{2}{c}{LV}
& \multicolumn{2}{c}{LA}
& \multicolumn{2}{c}{RA}
& \multicolumn{2}{c}{RV}
& \multicolumn{2}{c}{Average} \\
&
& Dice$\uparrow$ & NSD$\uparrow$
& Dice$\uparrow$ & NSD$\uparrow$
& Dice$\uparrow$ & NSD$\uparrow$
& Dice$\uparrow$ & NSD$\uparrow$
& Dice$\uparrow$ & NSD$\uparrow$
& Dice$\uparrow$ & NSD$\uparrow$ \\
\midrule
\midrule

\multicolumn{14}{@{}l}{\textit{\textbf{TS training set $\rightarrow$ TS test set}}} \\

\multirow{2}{*}{\textbf{(a)}}
& nnU-Net~\cite{isensee2021nnu}
& 80.22 & 83.67 & 78.06 & 75.16 & 88.32 & 82.24 & 83.21 & 73.43 & \textbf{88.48} & 81.62 & 83.66 & 79.22 \\

& SwinUNETR~\cite{hatamizadeh2021swin}
& 80.36 & 83.71 & 67.69 & 58.31 & 87.66 & 79.67 & 82.94 & 72.04 & 88.13 & 79.27 & 81.36 & 74.60 \\

\multirow{1}{*}{\textbf{(b)}}
& Registration Atlas~\cite{ding2022cross}
& NaN & NaN & NaN & NaN & NaN & NaN & NaN & NaN & NaN & NaN & NaN & NaN \\

\multirow{1}{*}{\textbf{(c)}}
& Input-Cond Diffusion~\cite{wu2024medsegdiff}
& 77.80 & 84.30 & 74.20 & 73.34 & 87.44 & 81.64 & 82.72 & 76.25 & 86.53 & 79.68 & 81.74 & 79.04 \\

\midrule

\multirow{2}{*}{\textbf{(d)}}
& \acronN{} (Ours)
& 83.52 & 88.87 & 77.21 & 75.70 & 90.13 & 86.35 & \textbf{86.52} & 80.76 & 88.36 & 82.06 & 85.17 & 82.74 \\

& \acronF{} (Ours)
& \textbf{83.80} & \textbf{89.62} & \textbf{78.06} & \textbf{76.01} & \textbf{90.77} & \textbf{87.46} & 86.29 & \textbf{81.37} & 88.27 & \textbf{82.49} & \textbf{85.44} & \textbf{83.39} \\

\midrule
\midrule

\multicolumn{14}{@{}l}{\textit{\textbf{CT training set $\rightarrow$ CT test set}}} \\

\multirow{2}{*}{\textbf{(a)}}
& nnU-Net~\cite{isensee2021nnu}
& 57.88 & 47.18 & 60.05 & 22.41 & 73.31 & 42.62 & 49.68 & 25.35 & 69.24 & 43.25 & 62.03 & 36.16 \\

& SwinUNETR~\cite{hatamizadeh2021swin}
& 54.22 & 43.67 & 60.63 & 22.05 & 67.59 & 32.77 & 47.46 & 20.15 & 64.21 & 34.99 & 58.82 & 30.73 \\

\multirow{1}{*}{\textbf{(b)}}
& Registration Atlas~\cite{ding2022cross}
& 51.13 & 40.56 & 65.28 & 30.66 & 60.04 & 31.13 & 66.96 & 39.54 & 55.38 & 29.47 & 59.76 & 34.27 \\

\multirow{1}{*}{\textbf{(c)}}
& Input-Cond Diffusion~\cite{wu2024medsegdiff}
& 58.03 & 52.63 & 41.98 & 26.05 & 68.51 & 45.64 & 66.47 & 41.81 & 39.98 & 25.56 & 54.99 & 38.34 \\

\midrule

\multirow{2}{*}{\textbf{(d)}}
& \acronN{} (Ours)
& 78.02 & 73.75 & 87.27 & 74.74 & 86.68 & 74.41 & \textbf{80.25} & \textbf{65.87} & 84.02 & \textbf{66.79} & 83.25 & 71.11 \\

& \acronF{} (Ours)
& \textbf{78.76} & \textbf{74.14} & \textbf{88.16} & \textbf{77.20} & \textbf{87.31} & \textbf{74.89} & 79.97 & 65.46 & \textbf{84.83} & 65.46 & \textbf{83.81} & \textbf{71.43} \\

\midrule
\midrule

\multicolumn{14}{@{}l}{\textit{\textbf{MRI training set $\rightarrow$ MRI test set}}} \\

\multirow{2}{*}{\textbf{(a)}}
& nnU-Net~\cite{isensee2021nnu}
& 41.11 & 38.90 & 51.63 & 20.28 & 67.63 & 31.98 & 49.66 & 22.00 & 50.05 & 28.99 & 52.02 & 28.43 \\

& SwinUNETR~\cite{hatamizadeh2021swin}
& 23.64 & 30.88 & 51.18 & 21.15 & 48.39 & 16.74 & 48.01 & 22.33 & 38.05 & 17.09 & 41.85 & 21.64 \\

\multirow{1}{*}{\textbf{(b)}}
& Registration Atlas~\cite{ding2022cross}
& 38.28 & 42.01 & 53.10 & 30.49 & 63.19 & 34.32 & 59.34 & 28.18 & 49.76 & 26.84 & 52.73 & 32.37 \\

\multirow{1}{*}{\textbf{(c)}}
& Input-Cond Diffusion~\cite{wu2024medsegdiff}
& 53.14 & 55.40 & 32.97 & 21.98 & 65.51 & 40.97 & 61.84 & 37.48 & 39.87 & 27.10 & 50.67 & 36.59 \\

\midrule

\multirow{2}{*}{\textbf{(d)}}
& \acronN{} (Ours)
& 57.97 & 46.42 & 70.67 & 41.11 & 75.70 & 43.93 & 73.11 & 44.14 & 65.77 & 41.64 & 68.64 & 43.45 \\

& \acronF{} (Ours)
& \textbf{56.63} & \textbf{36.36} & \textbf{72.56} & \textbf{40.19} & \textbf{74.99} & \textbf{46.25} & \textbf{76.01} & \textbf{50.01} & \textbf{67.53} & \textbf{44.63} & \textbf{69.54} & \textbf{43.49} \\

\midrule
\bottomrule
\end{tabular*}
}
\end{table*}

\sparag{Single-Modality: Table~\ref{tab:seg_normal}.}
We train and test using images acquired using the same modalities on the TS and MM-WHS datasets. In all cases. we use 80\% of the data for training and the remaining 20\% for testing. CMMAS~\cite{ding2022cross}, the registration-based atlas method, is limited by the high time and space complexity of the atlas paradigm, which restricts its scalability. As a result, it could not be executed on the large-scale TS dataset, hence the `NaN' in the table. 


\begin{table*}[htbp]
\caption{\textbf{Zero-shot cross-domain segmentation.} } 
\label{tab:seg_crossdomain}
\centering
{\scriptsize
\begin{tabular*}{\hsize}{@{}@{\extracolsep{\fill}}clcccccccccccc@{}}
\toprule
\midrule
\multirow{2}{*}{} & \multirow{2}{*}{Methods} & \multicolumn{2}{c}{Myo} & \multicolumn{2}{c}{LV} & \multicolumn{2}{c}{LA} & \multicolumn{2}{c}{RA} & \multicolumn{2}{c}{RV} & \multicolumn{2}{c}{Average} \\
& & Dice$\uparrow$ & NSD$\uparrow$ & Dice$\uparrow$ & NSD$\uparrow$ & Dice$\uparrow$ & NSD$\uparrow$ & Dice$\uparrow$ & NSD$\uparrow$ & Dice$\uparrow$ & NSD$\uparrow$ & Dice$\uparrow$ & NSD$\uparrow$ \\
\midrule
\midrule
\multicolumn{14}{@{}l}{\textit{\textbf{Cross-domain: CT training set $\rightarrow$ MRI test set}}} \\
\multirow{2}{*}{\textbf{(a)}}
& nnU-Net~\cite{isensee2021nnu} & 26.89 & 28.38 & 45.95 & 19.39 & 41.96 & 19.34 & 31.33 & 13.69 & 30.18 & 16.34 & 35.26 & 19.42 \\
& SwinUNETR~\cite{hatamizadeh2021swin} & 22.49 & 25.33 & 37.41 & 16.02 & 40.63 & 17.86 & 31.00 & 13.26 & 30.02 & 15.32 & 32.31 & 17.56 \\
\multirow{1}{*}{\textbf{(b)}}
& Registration Atlas~\cite{ding2022cross} & 31.70 & 34.53 & 42.01 & 24.88 & 50.71 & 27.71 & 43.05 & 29.92 & 39.72 & 23.55 & 41.44 & 28.12 \\
\multirow{1}{*}{\textbf{(c)}}
& Input-Cond Diffusion~\cite{wu2024medsegdiff} & 14.09 & 14.56 & 48.60 & 37.04 & 41.35 & 22.60 & 38.93 & 27.47 & 16.55 & 16.82 & 31.90 & 23.70 \\
\midrule
\multirow{2}{*}{\textbf{(d)}}
& \acronN{} (Ours) & 46.43 & 24.80 & 61.59 & 30.55 & 70.42 & 38.13 & 69.14 & 33.14 & 51.79 & 21.08 & 59.22 & 29.54 \\
& \acronF{} (Ours) & \textbf{47.26} & \textbf{28.40} & \textbf{62.00} & \textbf{31.19} & \textbf{70.37} & \textbf{38.58} & \textbf{65.76} & \textbf{32.73} & \textbf{54.17} & \textbf{21.70} & \textbf{59.91} & \textbf{30.52} \\
\midrule
\midrule
\multicolumn{14}{@{}l}{\textit{\textbf{Cross-domain: MRI training set $\rightarrow$ CT test set}}} \\
\multirow{2}{*}{\textbf{(a)}}
& nnU-Net~\cite{isensee2021nnu} & 25.03 & 30.29 & 47.54 & 20.01 & 56.74 & 24.12 & 52.59 & 21.85 & 48.46 & 18.22 & 46.07 & 22.90 \\
& SwinUNETR~\cite{hatamizadeh2021swin} & 24.06 & 28.03 & 51.22 & 23.58 & 45.23 & 17.86 & 42.88 & 17.97 & 48.93 & 20.45 & 42.46 & 21.58 \\
\multirow{1}{*}{\textbf{(b)}}
& Registration Atlas~\cite{ding2022cross} & 32.44 & 32.61 & 51.43 & 23.30 & 52.76 & 23.18 & 51.26 & 26.25 & 40.43 & 23.06 & 45.67 & 25.68 \\
\multirow{1}{*}{\textbf{(c)}}
& Input-Cond Diffusion~\cite{wu2024medsegdiff} & 46.52 & 20.84 & 41.53 & 11.84 & 67.41 & 18.54 & 66.62 & 18.21 & 16.25 & 4.91 & 47.67 & 14.87 \\
\midrule
\multirow{2}{*}{\textbf{(d)}}
& \acronN{} (Ours) & 54.24 & 33.72 & 72.22 & 36.97 & 61.90 & 21.58 & 69.14 & 32.06 & 61.49 & 27.78 & 63.80 & 30.42 \\
& \acronF{} (Ours) & \textbf{54.08} & \textbf{34.15} & \textbf{73.43} & \textbf{40.40} & \textbf{61.65} & \textbf{21.13} & \textbf{69.27} & \textbf{32.34} & \textbf{61.81} & \textbf{28.71} & \textbf{64.05} & \textbf{31.34} \\
\midrule
\bottomrule
\end{tabular*}
}
\end{table*}

\sparag{Zero-Shot Cross-Modality: Table~\ref{tab:seg_crossdomain}.}
We evaluate zero-shot cross-domain generalization on MM-WHS~\cite{zhuang2019evaluation} by training on one modality, CT or MRI, and testing on the other. Interestingly, since CT and MRI share similar anatomical structures, our method can generalize to unseen modalities  effectively adapting to new images at test time. In contrast, registration atlas methods lack such test-time adaptability, making it difficult to transfer atlas knowledge across modalities.


\begin{table*}[tb]
\caption{\textbf{Few-shot segmentation.}}
\label{tab:few_seg}
\centering
{\scriptsize
\begin{tabular*}{\textwidth}{@{}@{\extracolsep{\fill}}clcccccccccccc@{}}
\toprule
\midrule
\multirow{2}{*}{} & \multirow{2}{*}{Methods} & \multicolumn{2}{c}{Myo} & \multicolumn{2}{c}{LV} & \multicolumn{2}{c}{LA} & \multicolumn{2}{c}{RA} & \multicolumn{2}{c}{RV} & \multicolumn{2}{c}{Average} \\
& & Dice$\uparrow$ & NSD$\uparrow$ & Dice$\uparrow$ & NSD$\uparrow$ & Dice$\uparrow$ & NSD$\uparrow$ & Dice$\uparrow$ & NSD$\uparrow$ & Dice$\uparrow$ & NSD$\uparrow$ & Dice$\uparrow$ & NSD$\uparrow$ \\
\midrule
\midrule
\multicolumn{14}{@{}l}{\textit{\textbf{2-shot CT training set $\rightarrow$ CT test set}}} \\
\multirow{2}{*}{\textbf{(a)}}
& nnU-Net~\cite{isensee2021nnu} & 26.09 & 33.54 & 49.66 & 16.93 & 69.70 & 42.22 & 39.16 & 21.94 & 50.17 & 15.83 & 46.96 & 26.09 \\
& SwinUNETR~\cite{hatamizadeh2021swin} & NaN & NaN & NaN & NaN & NaN & NaN & NaN & NaN & NaN & NaN & NaN & NaN \\
\multirow{1}{*}{\textbf{(b)}}
& Registration Atlas~\cite{ding2022cross} & 44.29 & 39.16 & 58.83 & 13.85 & 67.99 & 31.68 & 61.83 & 33.45 & 59.83 & 32.70 & 58.55 & 30.17 \\
\multirow{1}{*}{\textbf{(c)}}
& Input-Cond Diffusion~\cite{wu2024medsegdiff} & 38.26 & 39.59 & 38.31 & 24.22 & 49.05 & 33.38 & 50.91 & 25.70 & 29.31 & 19.08 & 41.17 & 28.39 \\
\midrule
\multirow{2}{*}{\textbf{(d)}}
& \acronN{} (Ours) & \textbf{70.30} & 58.45 & 80.86 & 59.60 & \textbf{82.39} & \textbf{45.30} & \textbf{78.01} & \textbf{58.42} & \textbf{77.08} & 55.88 & \textbf{77.73} & 55.53 \\
& \acronF{} (Ours) & 70.03 & \textbf{61.29} & \textbf{83.73} & \textbf{61.11} & 80.17 & 44.91 & 77.59 & 55.16 & 76.84 & \textbf{57.27} & 77.67 & \textbf{55.95} \\

\midrule
\midrule
\multicolumn{14}{@{}l}{\textit{\textbf{4-shot CT training set $\rightarrow$ CT test set}}} \\
\multirow{2}{*}{\textbf{(a)}}
& nnU-Net~\cite{isensee2021nnu} & 35.35 & 31.13 & 60.25 & 18.71 & 71.78 & 31.14 & 40.40 & 16.29 & 64.39 & 40.73 & 54.43 & 27.60 \\
& SwinUNETR~\cite{hatamizadeh2021swin} & NaN & NaN & NaN & NaN & NaN & NaN & NaN & NaN & NaN & NaN & NaN & NaN \\
\multirow{1}{*}{\textbf{(b)}}
& Registration Atlas~\cite{ding2022cross} & 49.61 & 41.69 & 61.03 & 34.53 & 71.21 & 35.72 & 58.01 & 39.12 & 65.95 & 33.88 & 61.16 & 36.99 \\
\multirow{1}{*}{\textbf{(c)}}
& Input-Cond Diffusion~\cite{wu2024medsegdiff} & 43.32 & 40.74 & 38.49 & 24.55 & 62.80 & 38.74 & 51.05 & 25.59 & 28.65 & 19.34 & 44.86 & 29.79 \\
\midrule
\multirow{2}{*}{\textbf{(d)}}
& \acronN{} (Ours) & \textbf{72.05} & \textbf{53.21} & \textbf{83.07} & 56.69 & \textbf{86.72} & 44.40 & 74.56 & 41.35 & 79.50 & 55.69 & 79.18 & 50.27 \\
& \acronF{} (Ours) & 70.36 & 51.66 & 82.95 & \textbf{61.74} & 86.06 & \textbf{54.39} & \textbf{75.57} & \textbf{49.22} & \textbf{81.42} & \textbf{59.38} & \textbf{79.27} & \textbf{55.28} \\

\midrule
\midrule
\multicolumn{14}{@{}l}{\textit{\textbf{2-shot MRI training set $\rightarrow$ MRI test set}}} \\
\multirow{2}{*}{\textbf{(a)}}
& nnU-Net~\cite{isensee2021nnu} & 25.95 & 27.15 & 23.63 & 13.68 & 35.33 & 16.32 & 37.52 & 18.47 & 13.16 & 12.73 & 27.12 & 17.67 \\
& SwinUNETR~\cite{hatamizadeh2021swin} & NaN & NaN & NaN & NaN & NaN & NaN & NaN & NaN & NaN & NaN & NaN & NaN \\
\multirow{1}{*}{\textbf{(b)}}
& Registration Atlas~\cite{ding2022cross} & 33.74 & 35.17 & 40.24 & 25.99 & 51.59 & 28.73 & 61.10 & 33.63 & 38.38 & 23.38 & 45.01 & 29.38 \\
\multirow{1}{*}{\textbf{(c)}}
& Input-Cond Diffusion~\cite{wu2024medsegdiff} & 45.39 & 44.72 & 7.90 & 6.97 & 61.35 & 36.69 & 56.13 & 33.71 & 22.43 & 17.91 & 38.64 & 28.00 \\
\midrule
\multirow{2}{*}{\textbf{(d)}}
& \acronN{} (Ours) & \textbf{49.56} & \textbf{42.06} & 57.48 & 24.71 & 61.05 & \textbf{28.52} & \textbf{64.59} & 30.19 & 51.26 & 34.55 & 56.79 & 32.01 \\
& \acronF{} (Ours) & 47.64 & 33.38 & \textbf{58.41} & \textbf{26.42} & \textbf{69.13} & 28.15 & 57.54 & \textbf{36.92} & \textbf{54.43} & \textbf{38.39} & \textbf{57.43} & \textbf{32.65} \\

\midrule
\midrule
\multicolumn{14}{@{}l}{\textit{\textbf{4-shot MRI training set $\rightarrow$ MRI test set}}} \\
\multirow{2}{*}{\textbf{(a)}}
& nnU-Net~\cite{isensee2021nnu} & 33.12 & 32.30 & 48.79 & 25.47 & 55.48 & 26.19 & 41.42 & 19.53 & 31.78 & 19.57 & 42.12 & 24.61 \\
& SwinUNETR~\cite{hatamizadeh2021swin} & NaN & NaN & NaN & NaN & NaN & NaN & NaN & NaN & NaN & NaN & NaN & NaN \\
\multirow{1}{*}{\textbf{(b)}}
& Registration Atlas~\cite{ding2022cross} & 47.88 & 43.62 & 49.24 & 24.56 & 59.99 & 30.42 & 65.89 & 35.18 & 49.78 & 28.41 & 54.56 & 32.44 \\
\multirow{1}{*}{\textbf{(c)}}
& Input-Cond Diffusion~\cite{wu2024medsegdiff} & 49.19 & 44.88 & 20.64 & 16.48 & 65.81 & 35.22 & 55.19 & 33.74 & 9.44 & 17.09 & 40.06 & 29.48 \\
\midrule
\multirow{2}{*}{\textbf{(d)}}
& \acronN{} (Ours) & 54.34 & 48.42 & 69.69 & 39.43 & 72.09 & 44.78 & \textbf{75.92} & \textbf{48.09} & \textbf{63.22} & \textbf{41.46} & 67.05 & 44.44 \\
& \acronF{} (Ours) & \textbf{55.07} & \textbf{53.76} & \textbf{72.53} & \textbf{40.32} & \textbf{73.67} & \textbf{46.64} & 73.76 & 46.61 & 61.11 & 40.84 & \textbf{67.23} & \textbf{45.64} \\
\midrule
\bottomrule
\end{tabular*}
}
\end{table*}

\sparag{Few-Shot: Table~\ref{tab:few_seg}.}
We evaluate the few-shot setting on MM-WHS~\cite{zhuang2019evaluation}. The test set remains the same as in the previous setting, comprising 20\% of the total data, while the training set uses either 2 samples (2-shot) or 4 samples (4-shot). 


\begin{figure}[htbp]
        \centering
	\includegraphics[width=1\linewidth]{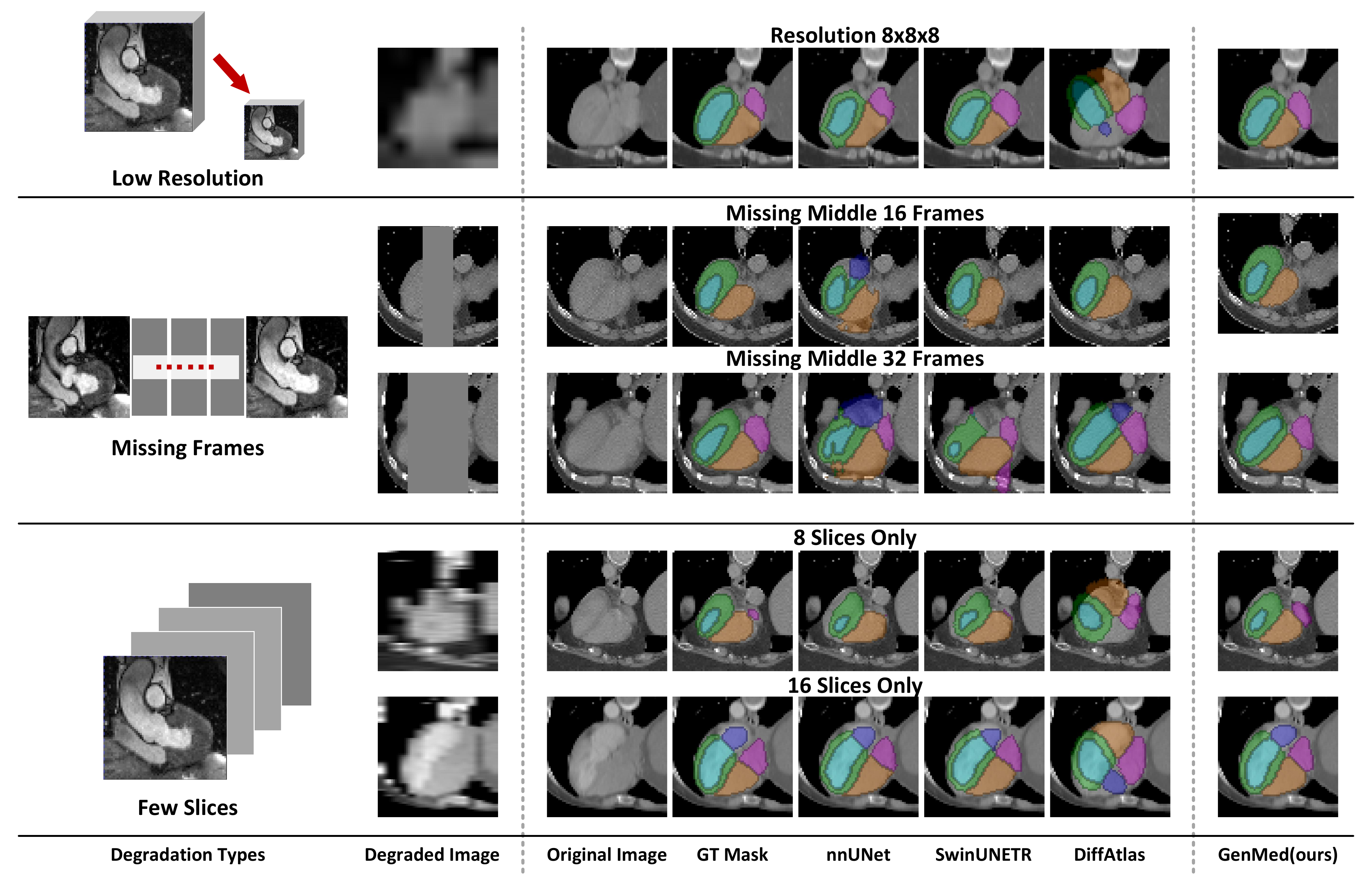}
	\vspace{-4mm}
        \caption{\textbf{Segmentation using degraded inputs.} \acronF{} is less affected than the baselines.}
	\label{fig:degrade}
\end{figure}

\begin{table*}[tb]
\caption{\textbf{Segmentation Using Degraded Inputs.}} 
\label{tab:seg_defects}
\centering
{\scriptsize
\begin{tabular*}{\textwidth}{@{}@{\extracolsep{\fill}}lcccccccccccc@{}}
\toprule
\midrule
\multirow{2}{*}{Methods} & \multicolumn{2}{c}{Myo} & \multicolumn{2}{c}{LV} & \multicolumn{2}{c}{LA} & \multicolumn{2}{c}{RA} & \multicolumn{2}{c}{RV} & \multicolumn{2}{c}{Average} \\
& Dice$\uparrow$ & NSD$\uparrow$ & Dice$\uparrow$ & NSD$\uparrow$ & Dice$\uparrow$ & NSD$\uparrow$ & Dice$\uparrow$ & NSD$\uparrow$ & Dice$\uparrow$ & NSD$\uparrow$ & Dice$\uparrow$ & NSD$\uparrow$ \\
\midrule
\midrule
\multicolumn{13}{@{}l}{\textit{\textbf{Resolution $8 \times 8 \times 8$}}} \\
nnU-Net~\cite{isensee2021nnu} & 57.09 & 52.03 & 58.50 & 39.07 & 74.14 & 42.88 & 68.82 & 44.48 & 77.85 & 53.33 & 67.28 & 46.36 \\
SwinUNETR~\cite{hatamizadeh2021swin} & 59.10 & 52.50 & 59.54 & 43.82 & 74.13 & 44.10 & 67.00 & 42.54 & 78.08 & 54.72 & 67.57 & 47.54 \\
\midrule
\acronN{} (Ours) & 18.54 & 20.51 & 1.93 & 1.22 & 24.21 & 11.54 & 26.86 & 15.06 & 4.70 & 3.89 & 15.25 & 10.44 \\
\acronF{} (Ours) & \textbf{69.17} & \textbf{64.99} & \textbf{66.44} & \textbf{50.21} & \textbf{81.93} & \textbf{58.80} & \textbf{77.06} & \textbf{57.76} & \textbf{81.09} & \textbf{60.32} & \textbf{75.14} & \textbf{58.42} \\
\midrule
\midrule
\multicolumn{13}{@{}l}{\textit{\textbf{Missing middle 8 frames}}} \\
nnU-Net~\cite{isensee2021nnu} & 75.76 & 78.32 & 61.40 & 50.04 & 84.86 & 73.99 & 83.89 & 74.67 & 83.41 & 68.84 & 77.87 & 69.17 \\
SwinUNETR~\cite{hatamizadeh2021swin} & 78.95 & 81.67 & 72.23 & 64.11 & 87.91 & 79.85 & 82.39 & 72.61 & 82.84 & 66.53 & 80.86 & 72.95 \\
\midrule
\acronN{} (Ours) & 82.41 & 87.14 & 75.51 & 71.03 & 89.74 & 85.08 & 85.83 & 79.08 & 87.70 & 79.71 & 84.24 & 80.41 \\
\acronF{} (Ours) & \textbf{84.05} & \textbf{89.63} & \textbf{76.59} & \textbf{73.42} & \textbf{90.86} & \textbf{87.87} & \textbf{86.62} & \textbf{82.25} & \textbf{88.11} & \textbf{81.16} & \textbf{85.25} & \textbf{82.87} \\
\midrule
\midrule
\multicolumn{13}{@{}l}{\textit{\textbf{Missing middle 16 frames}}} \\
nnU-Net~\cite{isensee2021nnu} & 67.82 & 69.87 & 40.85 & 25.29 & 81.05 & 66.10 & 81.35 & 69.06 & 70.56 & 45.15 & 68.33 & 55.09 \\
SwinUNETR~\cite{hatamizadeh2021swin} & 68.76 & 69.41 & 56.96 & 40.22 & 81.67 & 67.04 & 76.86 & 62.13 & 73.01 & 47.17 & 71.45 & 57.20 \\
\midrule
\acronN{} (Ours) & 79.50 & 82.50 & 71.26 & 60.88 & 87.72 & 80.07 & 84.33 & 75.73 & 84.91 & 72.47 & 81.54 & 74.33 \\
\acronF{} (Ours) & \textbf{83.62} & \textbf{88.98} & \textbf{76.93} & \textbf{73.92} & \textbf{90.51} & \textbf{87.34} & \textbf{84.32} & \textbf{77.37} & \textbf{85.84} & \textbf{75.41} & \textbf{84.24} & \textbf{80.60} \\
\midrule
\midrule
\multicolumn{13}{@{}l}{\textit{\textbf{Missing middle 32 frames}}} \\
nnU-Net~\cite{isensee2021nnu} & 43.68 & 42.33 & 27.37 & 9.33 & 62.14 & 38.92 & 67.48 & 46.83 & 52.51 & 21.88 & 50.64 & 31.86 \\
SwinUNETR~\cite{hatamizadeh2021swin} & 36.53 & 35.20 & 26.77 & 13.55 & 50.36 & 27.04 & 45.61 & 27.19 & 46.04 & 21.91 & 41.06 & 24.98 \\
\midrule
\acronN{} (Ours) & 59.52 & 60.07 & 53.59 & 37.04 & 71.06 & 55.33 & 68.52 & 53.24 & 69.10 & 48.20 & 64.36 & 50.78 \\
\acronF{} (Ours) & \textbf{67.35} & \textbf{67.15} & \textbf{57.24} & \textbf{40.21} & \textbf{78.29} & \textbf{61.64} & \textbf{76.74} & \textbf{64.01} & \textbf{73.51} & \textbf{52.04} & \textbf{70.63} & \textbf{57.01} \\
\midrule
\midrule
\multicolumn{13}{@{}l}{\textit{\textbf{8 Slices only}}} \\
nnU-Net~\cite{isensee2021nnu} & 35.60 & 35.38 & 56.49 & 36.14 & 60.44 & 21.24 & 58.75 & 26.15 & 69.45 & 37.07 & 56.15 & 31.20 \\
SwinUNETR~\cite{hatamizadeh2021swin} & 40.44 & 36.49 & 58.37 & 37.61 & 62.37 & 25.09 & 67.49 & 39.29 & 76.12 & 46.98 & 60.96 & 37.09 \\
\midrule
\acronN{} (Ours) & 34.51 & 35.45 & 19.32 & 13.17 & 45.34 & 25.86 & 45.56 & 26.27 & 26.69 & 18.03 & 34.29 & 23.76 \\
\acronF{} (Ours) & \textbf{53.29} & \textbf{41.58} & \textbf{61.92} & \textbf{39.53} & \textbf{73.38} & \textbf{33.49} & \textbf{68.62} & \textbf{33.84} & \textbf{79.29} & \textbf{53.58} & \textbf{67.30} & \textbf{40.40} \\
\midrule
\midrule
\multicolumn{13}{@{}l}{\textit{\textbf{16 Slices only}}} \\
nnU-Net~\cite{isensee2021nnu} & 65.59 & 57.70 & 71.32 & 60.88 & 80.23 & 51.45 & 77.19 & 54.17 & 83.30 & 65.73 & 75.53 & 57.98 \\
SwinUNETR~\cite{hatamizadeh2021swin} & 67.13 & 59.78 & 72.41 & 62.73 & 80.85 & 54.19 & 80.72 & 65.84 & 85.50 & 71.95 & 77.32 & 62.90 \\
\midrule
\acronN{} (Ours) & 61.75 & 62.99 & 51.29 & 43.20 & 71.31 & 57.93 & 69.32 & 53.77 & 60.54 & 50.61 & 62.84 & 53.70 \\
\acronF{} (Ours) & \textbf{76.68} & \textbf{78.73} & \textbf{73.51} & \textbf{66.30} & \textbf{86.82} & \textbf{74.61} & \textbf{82.77} & \textbf{70.88} & \textbf{87.01} & \textbf{76.94} & \textbf{81.36} & \textbf{73.49} \\
\midrule
\bottomrule
\end{tabular*}
}
\end{table*}

\sparag{Degraded Inputs: Table~\ref{tab:seg_defects}.}
To gauge sensitivity to input quality, we degrade the original high-resolution TS images in three different ways. We use low-resolution inputs of size $8\times8\times8$, remove 8, 16, or 32 middle slices, or retain only a small number of slices, as shown in Fig.~\ref{fig:degrade}.

\parag{Ablation Analysis.} For our \acronF{}, we adopt the smooth update formulation defined in Eq.~\ref{eq:softguidance_explicit}, in which the update strength is controlled by the hyperparameter $\eta$ and has a significant impact on model performance. To investigate this effect, we conduct an ablation study on different values of $\eta$ under the \textit{Missing middle 16 frames} setting from Table~\ref{tab:seg_defects}, with the resulting segmentation performance reported in Table~\ref{tab:seg_eta_ablation}. As shown, a small value (e.g., $\eta=0.5$) provides insufficient guidance for effective updates, whereas a large value (e.g., $\eta=5$) tends to destabilize the model and lead to performance degradation. Based on these results, we select $\eta=1.5$ as a suitable trade-off that ensures both stable and effective updates.

\begin{table*}[tb]
\caption{\textbf{Ablation results under different values of $\eta$ for defect segmentation.}}
\label{tab:seg_eta_ablation}
\centering
{\scriptsize
\begin{tabular*}{\textwidth}{@{}@{\extracolsep{\fill}}lcccccccccccc@{}}
\toprule
\multirow{2}{*}{$\eta$} & \multicolumn{2}{c}{Myo} & \multicolumn{2}{c}{LV} & \multicolumn{2}{c}{LA} & \multicolumn{2}{c}{RA} & \multicolumn{2}{c}{RV} & \multicolumn{2}{c}{Average} \\
& Dice$\uparrow$ & NSD$\uparrow$ & Dice$\uparrow$ & NSD$\uparrow$ & Dice$\uparrow$ & NSD$\uparrow$ & Dice$\uparrow$ & NSD$\uparrow$ & Dice$\uparrow$ & NSD$\uparrow$ & Dice$\uparrow$ & NSD$\uparrow$ \\
\midrule
0.5 & 81.97 & 85.97 & 72.88 & 63.90 & 89.58 & 83.68 & 60.13 & 79.77 & 86.22 & 75.56 & 78.15 & 77.77 \\
1.0 & 81.75 & 85.63 & 72.35 & 63.31 & 89.22 & 82.97 & 85.84 & 79.96 & 85.21 & 72.59 & 82.87 & 76.89 \\
1.5 & \textbf{83.62} & \textbf{88.98} & \textbf{76.93} & \textbf{73.92} & \textbf{90.51} & \textbf{87.34} & \textbf{84.32} & \textbf{77.37} & \textbf{85.84} & \textbf{75.41} & \textbf{84.24} & \textbf{80.60} \\
2.0 & 80.12 & 82.74 & 70.73 & 59.99 & 88.60 & 79.56 & 83.39 & 74.62 & 82.92 & 65.99 & 81.15 & 72.58 \\
5.0 & 63.23 & 39.72 & 58.60 & 32.08 & 83.53 & 38.18 & 58.69 & 34.65 & 79.29 & 54.67 & 68.66 & 39.86 \\

\bottomrule
\end{tabular*}
}
\end{table*}

\parag{Discussion.}  Under standard and few-shot settings, \acronN{} and \acronF{} outperform the other methods while performing similarly. However, in the more challenging scenarios, they also do better than the baselines but the more sophisticated guidance mechanism \acronF{} gives an edge over \acronN{}. An explanation is that, under easier conditions, the modeling of the joint distribution suffices to surpass the baselines while, under more difficult ones,  careful guidance is also required.

\subsection{Operating in Latent Space for Shape Completion}
\label{sec:exp_shape}

Unlike segmentation, 3D shape completion involves dense representations with prohibitive dimensionality, which necessitates the use of auxiliary encoders and decoders for compression and recovery to improve both robustness and computational efficiency. We therefore validate the latent space approach of Section~\ref{sec:latent} for the purpose of shape completion in a diverse set of medical 3D shapes~\cite{li2025medshapenet}. In this context, $X$ denotes the latent feature representing a partial shape encoded via a Signed Distance Function (SDF), $Y$ corresponds to the latent feature of the missing region, and $D^*$ represents the observed partial SDF. Because shapes are represented in the explicit space rather than the implicit one, we need to use a decoder, $D = \decd(X)$, to reconstruct the full SDF before computing the loss and back-propagating gradients using Eq.~\ref{eq:latent_loss}. 

\parag{Baselines.}
We benchmark representative open-source 3D shape diffusion frameworks spanning two paradigms: dense voxelised SDF on regular grids (SDFusion~\cite{cheng2023sdfusion}, Diffusion-SDF~\cite{li2023diffusion}) and octree-based adaptive latent representations (OctFusion~\cite{xiong2025octfusion}). Building upon these frameworks, we constructed strong baselines tailored for medical shape generation. Further details on shape representations and architecture are deferred to Appendix~\ref{sec:architecture}. Fixing the baseline architecture and representation, we then compare two families of guidance methods against ours:
\begin{itemize}
    \item \textit{Classifier-Free Guidance (CFG).} We consider two groups of variants: (i) different guidance schedules~\cite{xi2024analysis} -- Constant, Linear, and Exponential; and (ii) modifications to the CFG update rule, including APG~\cite{sadat2024eliminating}, Rectified-CFG++~\cite{sainirectified}, and CFG-Zero$^\star$~\cite{fan2025cfg}.
    \item \textit{Input Conditioning}. The conditioning signal is explicitly provided during training to directly model $P(Y \mid X)$, and is fed into the model as part of its input at inference time~\cite{zhang2026high,wang20253d}.
\end{itemize}

\parag{Metrics.}
For evaluation purposes, we adopt two classes of metrics. \emph{Shape generation quality} is measured by Minimum Matching Distance (MMD), Coverage (COV), and 1-Nearest-Neighbour Accuracy (1-NN), which together assess distributional fidelity and diversity. \emph{Shape completion quality} is measured by the Dice Coefficient (Dice), Chamfer Distance (CD), and Unidirectional Hausdorff Distance (UHD), capturing overlap, average geometric error, and worst-case deviation, respectively.

\parag{Datasets.}
We employ two datasets. The first evaluates baseline performance: the cleaned MedShapeNet~\cite{li2025medshapenet}, comprising 139 categories (22 instruments) and 68,754 samples (97 instruments), split 4:1 into 54,947 training and 13,807 testing samples. The second assesses zero-shot generalization on unseen, non-fine-tuned data: 100 cases of 3D eyeball shapes from the Fundus2Globe dataset~\cite{shi2025fundus2globe}, derived from the Zhongshan Ophthalmic Center–Brien Holden Vision Institute (ZOC-BHVI) Guangzhou High Myopia Cohort Study. Further details of both datasets are provided in Appendix~\ref{sec:shape_dataset}.

\begin{table}[htbp]
\centering
\caption{\textbf{Shape Generation and Different Guidance Strategies for Shape Completion under Diverse Prompts.} Group~1 evaluates multi-category shape generation on MedShapeNet; Group~2 compares different guidance strategies for shape completion under multi-plane prompts based on the best framework from Group~1; Group~3 focuses on shape completion across various corruption prompts (\% values).}
\label{tab:merged_compact_results_mixed}
\scriptsize
\setlength{\tabcolsep}{3pt}
\begin{tabular}{@{}llccc@{}}
\toprule
\textbf{Type} & \textbf{Method}
  & MMD$\downarrow$
  & COV$\uparrow$
  & 1-NN$\downarrow$ \\
\midrule
\multirow{4}{*}{Text-only}
  & SDFusion~\cite{cheng2023sdfusion}    & 2.07 & 52.33 & 70.44 \\
  & Diffusion-SDF~\cite{li2023diffusion} & 3.19 & 41.05 & 83.49 \\
  & OctFusion~\cite{xiong2025octfusion}  & 8.75 & 24.14 & 87.56 \\
  & \acronB{} (Ours)            & \textbf{1.16} & \textbf{52.80} & \textbf{66.24} \\
\midrule
\multicolumn{2}{@{}l}{\textbf{\textit{Comparison of Multi-Plane Prompt Generation Strategies}}}
  & Dice$\uparrow$ & CD$\downarrow$ & UHD$\downarrow$ \\
\cmidrule(lr){3-5}
\multirow{6}{*}{CFG}
  & Constant~\cite{xi2024analysis}                  & 61.19          & 0.67           & 13.92          \\
  & Linear Schedule~\cite{xi2024analysis}           & 55.33          & 1.94           & 19.02          \\
  & Exponential Schedule~\cite{xi2024analysis}      & 63.26          & 0.55           & 12.42          \\
  & APG~\cite{sadat2024eliminating}                 & 63.92          & 0.59           & 12.94          \\
  & Rectified-CFG++~\cite{sainirectified}                & 56.93          & 1.04           & 14.60          \\
  & CFG-Zero$^\star$~\cite{fan2025cfg}              & 54.44          & 2.15           & 17.96          \\
\cmidrule(lr){2-5}
  & Input Conditioning~\cite{zhang2026high,wang20253d} & 67.66          & \textbf{0.21}  & 12.32          \\
  & \acron{} (Ours)                                    & \textbf{75.94} & 0.75           & \textbf{10.16} \\
\midrule
\multicolumn{2}{@{}l}{\textbf{\textit{Shape Completion across Different Visual Prompts}}}
  & Dice$\uparrow$ & CD$\downarrow$ & UHD$\downarrow$ \\
\cmidrule(lr){3-5}
\multirow{5}{*}{\makecell[l]{Input Conditioning \\ \cite{zhang2026high,wang20253d}}}
  & Broken     & 69.22 & 0.21 & 12.98 \\
  & Oneplane   & 42.63 & 2.77 & 23.10 \\
  & Triplane   & 49.91 & 2.31 & 21.29 \\
  & Multiplane & 67.66 & 0.21 & 12.32 \\
  & Random     & 35.11 & 3.86 & 27.91 \\
\midrule
\multirow{5}{*}{ \acron{} (Ours)}
  & Broken     & 84.19 & 0.13 & 10.40 \\
  & Oneplane   & 51.18 & 2.08 & 21.11 \\
  & Triplane   & 69.74 & 1.02 & 14.01 \\
  & Multiplane & 75.94 & 0.75 & 10.16 \\
  & Random     & 53.26 & 1.61 & 18.87 \\
\bottomrule
\end{tabular}
\end{table}
\parag{Comparative Results.} We first compare various shape representations and architectures on medical shape generation under text-only conditioning to motivate our design, denoted \acronB{}. We then evaluate \acron{} on shape completion, comparing different Classifier-Free Guidance (CFG) strategies and input conditioning schemes across various settings.

\sparag{Shape Generation: Table~\ref{tab:merged_compact_results_mixed} Group 1.} 
The first group of text-only experiments demonstrates the architectural effectiveness of \acronB{} for medical shape generation. We adopt dense voxelised SDF on regular grids~\cite{cheng2023sdfusion,li2023diffusion} rather than octree-based adaptive latent representations~\cite{xiong2025octfusion}, and introduce architectural modifications tailored to the characteristics of medical shapes. This design yields shapes with higher fidelity to the ground truth, more comprehensively covers the diversity of real shape modes, and produces a generated distribution more consistent with the real medical shape distribution.

\sparag{Shape Completion: Table~\ref{tab:merged_compact_results_mixed} Group 2.}
Building upon the text conditioning setup, we further evaluate how different modeling paradigms affect guidance effectiveness. We first adopt the \textit{Multiplane} setting, in which only a limited number of slices are available as the prompt, and the model is required to recover the corresponding complete target shape. The results are reported in the second group of Table~\ref{tab:merged_compact_results_mixed}. An interesting observation is that, despite trying a range of CFG variants and performing a hyperparameter search (see Appendix~\ref{sec:search}), CFG-based guidance tends to underperform direct Input Conditioning in our setting. A possible explanation is that CFG was designed for weak, semantic-level conditions such as text, where amplifying the conditional direction helps alignment. Shape prompts like multi-plane masks instead impose dense, voxel-level geometric constraints that already tightly restrict the generation space, leaving little room for further guidance.
In this regime, extrapolation along the conditional direction may push samples off the geometric manifold, producing distortions that remain visually plausible but are penalised by Dice, CD, and UHD. In contrast, our \acron{} naturally adapts to such dense visual corruption prompts.

\begin{figure*}[h]
\centering
\includegraphics[width=1\linewidth]{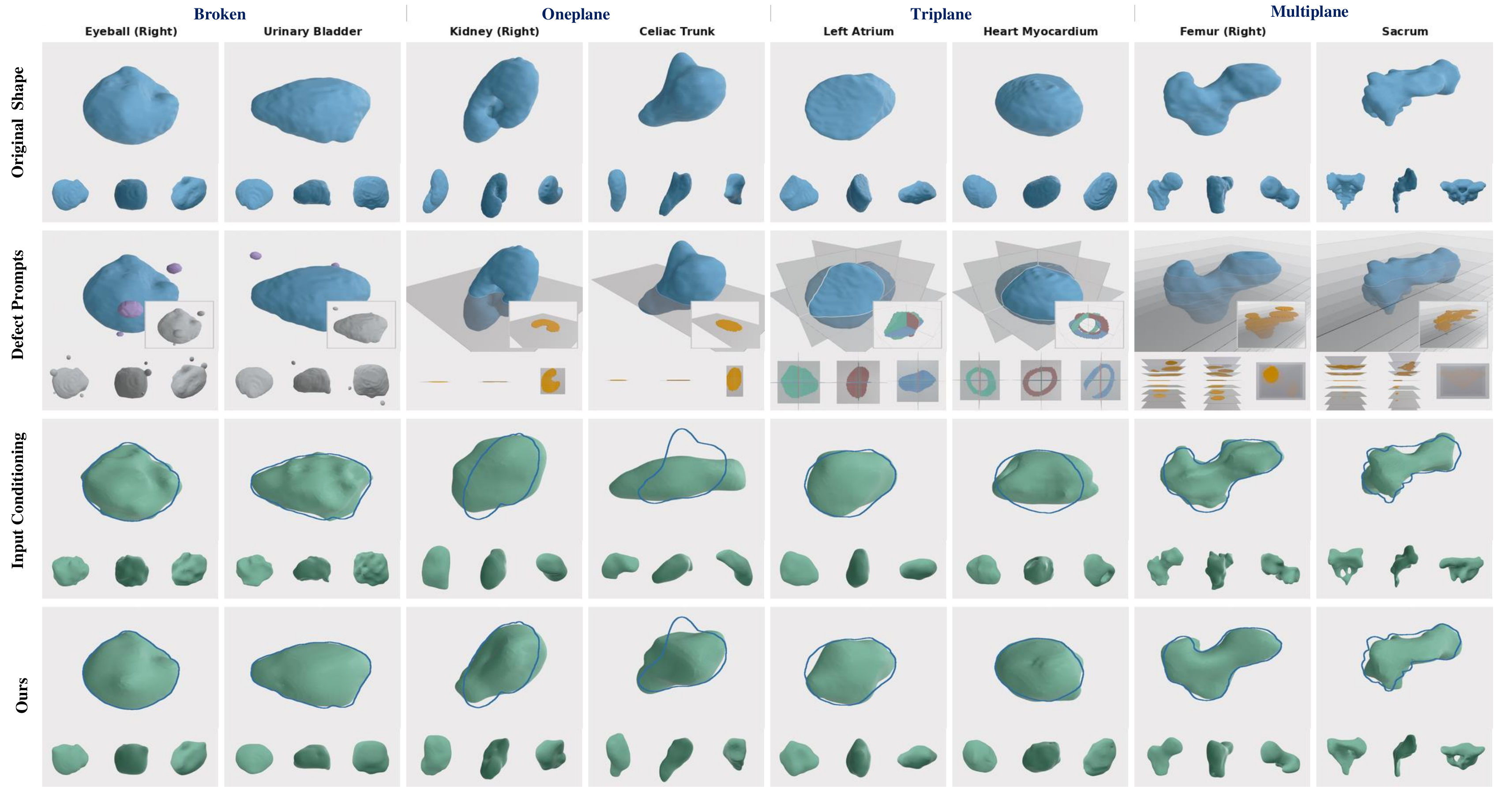}
\caption{\textbf{Visualization of shape completion under different visual prompts.}
Compared with input conditioning, \acron{} produces completions that
align more closely with the blue ground-truth boundaries across organs and
tissues of increasing structural complexity (e.g., eyeball, urinary bladder,
heart, and bone).}
\vspace{-10px}
\label{fig:shape_vis}
\end{figure*}

\sparag{Diverse Corruptions: Table~\ref{tab:merged_compact_results_mixed} Group 3.}
Building on the previous observation, the third group of
Table~\ref{tab:merged_compact_results_mixed} focuses on a direct comparison
between \acron{} and Input Conditioning under a range of clinically
motivated incomplete inputs. In real clinical scenarios, data corruption is
often unpredictable and may arise from factors such as acquisition devices
or storage issues, leading to diverse and stochastic degradation patterns.
To reflect this, we consider \textit{Broken} (randomly missing regions) as
well as more extreme cases in which only a limited number of slices are
available, namely \textit{Oneplane}, \textit{Triplane}, and
\textit{Multiplane}. Moreover, the available prompt is not restricted to a
single form and may consist of several types combined together. To capture
such realistic conditions, we additionally evaluate a \textit{Random}
setting, in which prompt combinations are sampled stochastically across the
different corruption types. The results show that \acron{} learns an
effective shape prior and remains robust across diverse and stochastic
corruption prompts, including the highly challenging \textit{Oneplane}
setting where only a single slice is available. Fig.~\ref{fig:shape_vis} visualises shape completion across organs of
increasing structural complexity, covering \textit{Broken} shapes and
limited-frame prompts (\textit{Oneplane}, \textit{Triplane}, and
\textit{Multiplane}). The first row shows the original shapes, used as
ground truth for the metrics in
Table~\ref{tab:merged_compact_results_mixed}. The second row illustrates
the corruption patterns: for \textit{Broken}, clinical device errors or
segmentation artefacts may cause missing or redundant regions (highlighted
in purple); the remaining columns show frame-based prompts as small front,
left, and top views. The third and fourth rows give the completion results
of Input Conditioning and \acron{} (Ours). Under \textit{Broken}, both
methods largely preserve the global structure thanks to learned shape
priors, but \acron{} recovers finer boundary details that align more
closely with the blue ground truth. Under limited-frame prompts, Input
Conditioning struggles to cover all corruption patterns at training time
and often fails to match the expected contours; in contrast, \acron{}
consistently generates shapes that adhere closely to the desired
completion boundaries.

\begin{figure*}[h]
    \centering
    \includegraphics[width=\linewidth]{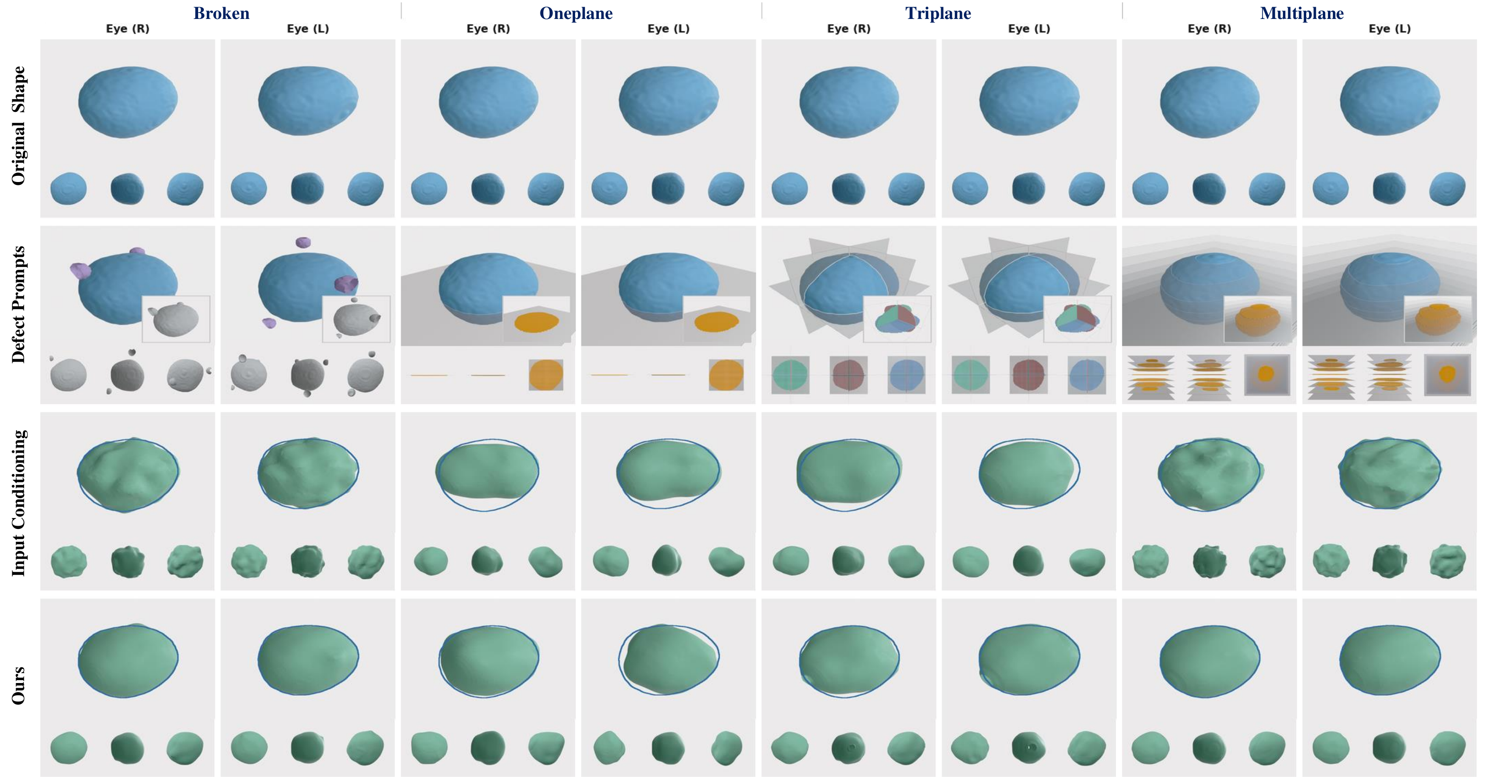}
   \caption{\textbf{Zero-shot shape completion on 3D eyeball under different
defect prompts.} \acron{} aligns more closely with the blue ground truth
than Input Conditioning.}
    \vspace{-10px}
    \label{fig:zero_eyes}
\end{figure*}

\sparag{Zero-Shot Shape Completion: Table~\ref{tab:eyeball_results}.} 
We further conduct zero-shot generation experiments on a new domain.
As shown in Table~\ref{tab:eyeball_results}, by applying control
optimisation at the output stage, \acron{} reduces the reliance on
large-scale training data and consistently outperforms the Input
Conditioning baseline. Fig.~\ref{fig:zero_eyes} visualises the inference
results on a new eyeball dataset without any fine-tuning, where we observe
trends consistent with Fig.~\ref{fig:shape_vis}: under \textit{Broken},
where only coarse structural cues are provided, \acron{} produces shapes
that more closely match the ground truth, and under different frame-based
prompts it likewise yields more stable and accurate completions.
\begin{table}[htbp]
\centering
\caption{\textbf{Zero-Shot of Different Guidance Strategies under Diverse Prompts on Eyeball Data (\% values).}}
\label{tab:eyeball_results}
\vspace{2pt}
\scriptsize
\setlength{\tabcolsep}{2.8pt}
\renewcommand{\arraystretch}{0.95}
\resizebox{\columnwidth}{!}{%
\begin{tabular}{llccccccccc}
\toprule
\multirow{2}{*}{\textbf{Methods}} & \multirow{2}{*}{\textbf{Setting}} 
& \multicolumn{3}{c}{\textbf{Eyeball Left}} 
& \multicolumn{3}{c}{\textbf{Eyeball Right}} 
& \multicolumn{3}{c}{\textbf{Average}} \\
\cmidrule(lr){3-5} \cmidrule(lr){6-8} \cmidrule(lr){9-11}
&  & Dice$\uparrow$ & CD$\downarrow$ & UHD$\downarrow$
   & Dice$\uparrow$ & CD$\downarrow$ & UHD$\downarrow$
   & Dice$\uparrow$ & CD$\downarrow$ & UHD$\downarrow$ \\
\midrule
\multirow{4}{*}{\shortstack[l]{Input\\Conditioning}}
& Oneplane   & 83.76 & 0.84 & 15.59 & 82.81 & 0.89 & 15.84 & 83.29 & 0.87 & 15.71 \\
& Triplane   & 86.41 & 0.65 & 14.50 & 87.06 & 0.62 & 14.08 & 86.74 & 0.64 & 14.29 \\
& Multiplane & 94.06 & 0.27 & 14.60 & 93.97 & 0.28 & 13.20 & 94.02 & 0.28 & 13.90 \\
& Broken     & 94.57 & 0.28 & 16.15 & 94.55 & 0.28 & 15.35 & 94.56 & 0.28 & 15.75 \\
\midrule
\multirow{4}{*}{\shortstack[l]{\acron{}\\(Ours)}}
& Oneplane   & 87.55 & 0.68 & 14.25 & 87.84 & 0.66 & 15.03 & 87.70 & 0.67 & 14.64 \\
& Triplane   & 95.65 & 0.26 & 9.24  & 95.45 & 0.26 & 9.66  & 95.55 & 0.26 & 9.45 \\
& Multiplane & 98.15 & 0.19 & 6.60  & 98.21 & 0.19 & 6.45  & 98.18 & 0.19 & 6.53 \\
& Broken     & 98.13 & 0.20 & 10.15 & 98.17 & 0.19 & 8.76  & 98.15 & 0.19 & 9.45 \\
\bottomrule
\end{tabular}%
}
\vspace{-4pt}
\end{table}

\definecolor{lightblue}{RGB}{220,230,255}
\begin{table}[htbp]
\centering
\caption{\textbf{Frequency and Scale for Shape Completion.}}
\label{tab:ablation_shape}
 \footnotesize
\setlength{\tabcolsep}{12pt}
\renewcommand{\arraystretch}{1.0}
\begin{tabular}{@{}cccc@{}}
\toprule
Scale 
& Dice$\uparrow$ 
& CD$\downarrow$ 
& UHD$\downarrow$ \\
\midrule
\multicolumn{4}{c}{\textit{Interval = 1, Time = 4.00 s}} \\
0.5 & 75.81 & 0.75 & 9.96 \\
0.7 & 80.82 & 0.49 & \textbf{8.51} \\
1.0 & 83.18 & 0.33 & 8.58 \\
1.3 & 83.61 & 0.25 & 8.97 \\
\rowcolor{lightblue}
1.5 & \textbf{83.65} & 0.23 & 9.34 \\
2.0 & 83.09 & \textbf{0.21} & 10.17 \\
5.0 & 79.12 & \textbf{0.21} & 15.19 \\
\midrule
\multicolumn{4}{c}{\textit{Interval = 2, Time = 2.73 s}} \\
0.5 & 57.79 & 1.68 & 16.60 \\
0.7 & 64.98 & 1.37 & 14.13 \\
1.0 & 72.67 & 0.95 & 11.30 \\
1.3 & 76.94 & 0.79 & 9.84 \\
1.5 & 78.72 & 0.66 & 9.29 \\
2.0 & \textbf{80.21} & 0.57 & \textbf{9.22} \\
5.0 & 76.98 & \textbf{0.37} & 11.93 \\
\midrule
\multicolumn{4}{c}{\textit{Interval = 4, Time = 1.57 s}} \\
0.5 & 45.20 & 2.56 & 21.43 \\
0.7 & 48.70 & 2.43 & 20.40 \\
1.0 & 53.47 & 2.18 & 18.86 \\
1.3 & 57.57 & 1.96 & 17.49 \\
1.5 & 60.02 & 1.85 & 16.80 \\
2.0 & 64.80 & 1.65 & 15.17 \\
5.0 & \textbf{70.57} & \textbf{1.26} & \textbf{13.14} \\
\bottomrule
\end{tabular}
\end{table}
\parag{Ablation Analysis.} In this section, we ablate the key hyperparameters of our framework. We adopt DDIM sampling with 20 steps to achieve efficient inference. We study the update strength $\eta$ in Eq.~\ref{eq:softguidance} and the guidance interval, where the number of effective guidance applications equals the total number of DDIM steps divided by the interval. As shown in Tab.~\ref{tab:ablation_shape}, increasing the interval reduces the number of guidance updates, lowering runtime at the cost of reconstruction quality. Specifically, as the interval increases from 1 to 4, the runtime drops from 4.00\,s to 1.57\,s, while the best Dice score falls from 83.65 to 70.57, and distance-based metrics degrade accordingly (best CD from 0.21 to 1.26; best UHD from $\sim$8.5 to 13.14). These results reveal a clear trade-off between efficiency and performance. We also observe that the optimal $\eta$ is strongly coupled with the interval: as guidance is applied less frequently, a larger update strength is needed to compensate, with the optimal $\eta$ increasing from 1.5 at interval $=1$ to 2.0 at interval $=2$, and further to 5.0 at interval $=4$. We further introduce a recurrent guidance mechanism~\cite{zhang2024lefusion,Lugmayr22}, in which the denoising trajectory is repeated multiple times at each guided timestep before advancing to the next step. As shown in Tab.~\ref{tab:ablation_recurrent}, we select recurrent $= 2$ as a favorable trade-off between runtime and performance. It is worth noting that all guidance strategies compared in this section share the same model architecture and thus incur identical memory costs. Further implementation details are provided in Appendix~Sec.~\ref{sec:shape_completetion}.
\begin{table}[htbp]
\centering
\caption{\textbf{Recurrent Steps for Shape Completion.}}
\label{tab:ablation_recurrent}
\footnotesize
\setlength{\tabcolsep}{12pt}
\renewcommand{\arraystretch}{1.0}
\begin{tabular}{@{}ccccc@{}}
\toprule
Recurrent
& Time (s)
& Dice$\uparrow$
& CD$\downarrow$
& UHD$\downarrow$ \\
\midrule
1 & 4.00  & 83.65          & 0.23          & \textbf{9.34}  \\
\rowcolor{lightblue}
2 & 6.00  & 84.19          & \textbf{0.13} & 10.40          \\
4 & 12.63 & \textbf{84.57} & 0.15          & 11.80          \\
6 & 19.08 & 84.55          & 0.21          & 13.93          \\
\bottomrule
\end{tabular}
\end{table}

\parag{Discussion.}
We compared the approach of \acron{} against classifier free guidance and input guidance. We also attempted to benchmark alternative loss-guidance paradigms, but found that these methods are tightly coupled to specific diffusion formulations and architectural choices, making a fair comparison difficult. For example, we tried adapting D-Flow~\cite{ben2024d} to the shape completion task. However, its requirement of LBFGS-based optimization over the initial noise made it computationally intractable within a practical wall-clock budget. Moreover, the cosine schedule and $\varepsilon$-prediction parameterization in our framework caused the back-propagation chain to become numerically unstable, leading directly to divergence. Despite the technical differences among these guidance approaches, they all steer the \emph{output} generation conditioned on a fixed \emph{input} signal, leaving the model's fundamental input-output contract unchanged. We view such loss-guidance designs as orthogonal and complementary to our pairwise modeling of $P(X,Y)$, rather than competing with it; a stronger guidance scheme would inherently translate to improved performance for our framework as well. We therefore do not include a direct empirical comparison against this class of methods.

\begin{figure*}[h]
\centering
\includegraphics[width=1\linewidth]{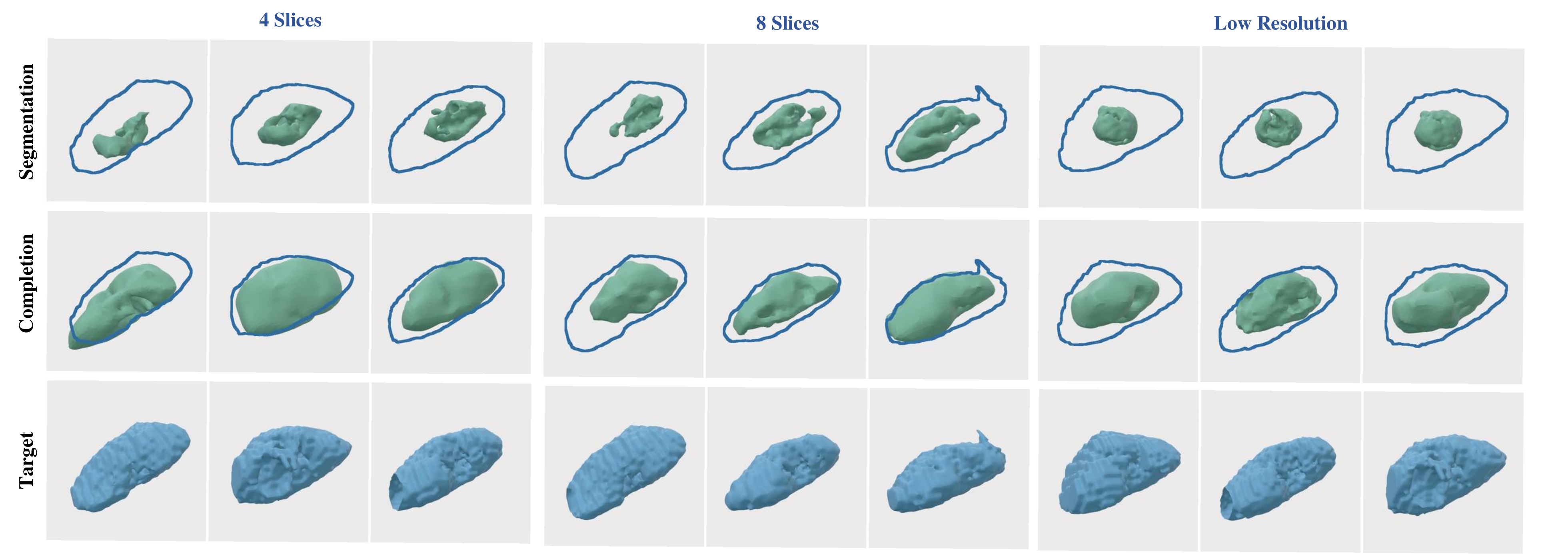}
\caption{\textbf{Visualization of Different Stages under Different Degradations.}
Rows 1-2 (green) show the segmentation results and the refined outputs after shape completion; row 3 (blue) shows the corresponding ground-truth shapes.}
\vspace{-10px}
\label{fig:diag_vis}
\end{figure*}

\subsection{Joint Framework for Diagnostic Applications}
\label{sec:diagnostic_framework}
The proposed paradigm \acron{} supports both image segmentation and shape completion. We integrate these two tasks into a single pipeline, enabling end-to-end transformation from raw imaging data to the final reconstructed shape. This demonstrates the applicability of our framework to clinical diagnostic workflows. Built upon the unified pairwise modeling paradigm in Sec.~\ref{sec:pairwise}, the framework adopts a two-stage design. In the first stage, the segmentation model from Sec.~\ref{sec:improved} converts CT scans into volumetric masks, which remains robust to corrupted or incomplete imaging data. In the second stage, shape completion is performed in the latent space (Sec.~\ref{sec:latent}) to recover the complete anatomical structure from the segmented shape.

\parag{Datasets.} Our pipeline consists of two stages: segmentation and shape completion. For evaluation, we use KiTS23~\cite{heller2023kits21}, which contains 599 contrast-enhanced abdominal CT scans with semantic annotations, from which we extract the left kidney label of each case. CT intensities are clipped to $[-250, 450]$ HU and rescaled to $[-1, 1]$. To better reflect real clinical scenarios, we evaluate under degraded CT inputs covering three challenging cases: highly incomplete volumes containing only 4 slices, less severely corrupted volumes containing 8 slices, and a low-resolution setting in which the input is resampled from $64 \times 64 \times 64$ voxels to $8 \times 8 \times 8$ voxels. The 489 training cases are randomly split (seed = 23) into 391 training, 49 validation, and 49 test cases. For the shape completion stage, we directly adopt the pre-trained shape priors from MedShapeNet~\cite{li2025medshapenet} described in Sec.~\ref{sec:exp_shape}. Since KiTS23 is independent of the dataset used to train the shape completion model, data leakage is avoided. Further implementation details can be found in Appendix Sec.~\ref{sec:diag_details}.

\begin{table}[t]
\centering
\caption{\textbf{Effect of Shape Completion under Different Degradations.} We report the Dice (\%) for segmentation and shape completion stages}
\label{tab:degradation_results}
\small
\setlength{\tabcolsep}{6pt}
\begin{tabular}{lccc}
\toprule
Degradation & Segmentation & Shape Completion & Gain \\
\midrule
4 Slices       & 50.22 & 64.16 & +13.94 \\
8 Slices       & 65.49 & 69.50 & +4.01 \\
Low Resolution & 38.95 & 50.11 & +11.16 \\
\bottomrule
\end{tabular}
\end{table}

\parag{Results.} Table~\ref{tab:degradation_results} illustrates the respective roles and contributions of the two stages (segmentation and shape completion.) in our framework for diagnostic applications. Even under the highly challenging \emph{4 Slices} setting, the first-stage segmentation model still achieves a Dice score of 50.22\%, indicating that it can provide a reasonable coarse estimate from extremely limited input data. At the same time, the second-stage shape completion is an essential component of the overall pipeline. For example, in the \emph{Low Resolution} setting, although the segmentation result is still limited in accuracy, applying shape completion with our framework and the pretrained shape prior leads to a substantial improvement, yielding a gain of 11.16 Dice points. More broadly, the results suggest that for more challenging degradation settings, such as \emph{4 Slices} and \emph{Low Resolution}, the first-stage segmentation performance tends to deteriorate more noticeably, while the subsequent shape completion stage provides larger improvements by leveraging the pretrained shape prior. Fig.~\ref{fig:diag_vis} visualizes the two-stage diagnostic pipeline in our unified framework under degraded CT inputs. As shown, under severe degradation settings such as \emph{4 Slices} and \emph{Low Resolution}, the segmentation stage can only recover the anatomically visible regions from the limited imaging evidence, often producing underestimated shapes and failing to preserve the full organ geometry. In contrast, the second-stage shape completion leverages the pretrained shape prior to effectively correct these incomplete segmentation results, compensating for the lack of global shape awareness caused by corrupted or insufficient observations.

\section{Conclusion}

We presented \emph{GenMed}, a generative reformulation of discriminative medical tasks. By modeling the joint distribution $p(X, Y)$ through a diffusion-based framework and applying test-time output optimization, GenMed shifts the paradigm from input conditioning to output-level guidance. This enables flexible, training-free adaptation to real-world clinical scenarios involving partial, noisy, low-resolution, or modality-shifted inputs---settings where traditional discriminative or prompt-based models often fail, especially in the absence of extensive data and augmentation. Our experiments demonstrate strong performance on standard settings and highlight GenMed's unique ability to generalize across tasks and observation types without task-specific retraining. Furthermore, we establish its utility as a general-purpose medical shape prior, capable of unified shape completion from diverse and even unseen observation forms. Beyond its practical utility, our generative perspective enables models to treat data not as static inputs but as partial observations of an underlying reality. We believe this shift opens promising avenues for more flexible and generalizable medical AI systems, and the framework readily extends to other domains, as demonstrated in Appendix Sec.~\ref{sec:generalization}.
\bibliographystyle{IEEEtran}     
\bibliography{cvlab-bib/string,cvlab-bib/vision,cvlab-bib/learning,cvlab-bib/biomed,cvlab-bib/reference}

\appendices
 \newpage
\appendix
\section{Technical Appendices}

\subsection{Segmentation}
\label{sec:seg_details}
For the segmentation experiments, we follow the experimental setup from our conference version~\cite{zhang2025diffatlas}, using the Adam optimizer with a learning rate of \(1 \times 10^{-4}\) (\(\beta_1 = 0.9\), \(\beta_2 = 0.999\)) and the L1 loss as the objective function. For the 2-shot setting, the batch size is set to 1 with approximately 10,000 training steps; for the 4-shot setting, the batch size is increased to 2 with roughly 15,000 steps; for all other experiments, the batch size is set to 12 and training proceeds for approximately 25,000 steps.

\subsection{Shape Completion}
\label{sec:shape_completetion}

\subsubsection{Dataset Construction}
\label{sec:shape_dataset}

We construct a 3D shape dataset of anatomical structures and medical instruments by filtering MedShapeNet. The dataset is designed to support both model training and benchmarking; therefore, we adopt widely used dataset organization and normalization practices from recent 3D shape studies, including 3DShape2VecSet and DORA.

The initial collection contains 106,101 meshes spanning 303 original category names. Since category names in MedShapeNet may include synonyms, naming variations, or definitions at different levels of granularity, we harmonize semantically equivalent labels into a unified taxonomy; for example, ribs are separated into twelve anatomically corresponding left/right classes. After label harmonization and mesh-level filtering, the resulting dataset contains 68,754 3D shapes across 139 categories. Among them, the instrument category includes 22 classes with 97 samples in total, while the non-instrument category includes 117 classes with 68,657 samples. For data splitting, we adopt a stratified train/test partition. The training set contains 54,947 samples, including 67 instrument samples and 54,880 non-instrument samples. The test set contains 13,807 samples, including 30 instrument samples and 13,777 non-instrument samples. Additionally, for low-frequency categories, we ensure that the test set contains at least one sample to guarantee coverage and fairness in evaluation.

To standardize geometric inputs and improve reproducibility across categories, we apply a preprocessing pipeline consistent with 3DShape2VecSet and DORA. Specifically, we exclude meshes that cannot be parsed or do not form valid surfaces, retain the largest connected component when necessary, and normalize each mesh by centering and scaling it to fit within a unit cube. For methods based on implicit representations, we further provide signed distance fields (SDFs) computed from the normalized meshes. Moreover, each category is paired with a concise natural-language description (including its name, morphological characteristics, and typical anatomical location), along with a corresponding 5120-dimensional $\ell_2$-normalized text embedding to facilitate retrieval and conditional generation.

\textbf{Text--Shape Pair Construction.}
To construct a text-shape paired dataset, we associate each category with a predefined textual description obtained from a curated mapping. These descriptions are used as prompts to guide shape generation and enable the model to learn category-level geometric priors.

The basic procedure is as follows: we input each category in the dataset into ChatGPT-5.2 Thinking and let the model output a detailed description of this category. For organs, the specific instruction is: please output a detailed description of [CATEGORY], including the shape, anatomical location, and characteristics of the organ; for medical instruments, the specific instruction is: please output a detailed description of [CATEGORY], including the function and shape of the medical instrument. We respectively provide the basic example cases for two types of cases: organs and medical instruments.

\textcolor{teal}{\textbf{Organ Test Cases:}}

\textcolor{gray}{\textbf{Test Case 1:}}
\begin{itemize}
\item \textbf{Input:} \texttt{heart}
\item \textbf{Output:}
\textcolor{green!50!black}{Heart, a muscular organ in the mediastinum between the lungs, featuring four chambers (two atria, two ventricles), coronary vessels, and valves that pump blood through pulmonary and systemic circulations.}
\end{itemize}

\textcolor{gray}{\textbf{Test Case 2:}}
\begin{itemize}
\item \textbf{Input:} \texttt{kidney\_left}
\item \textbf{Output:}
\textcolor{green!50!black}{Left kidney, a bean-shaped retroperitoneal organ located at the T12-L3 level, featuring cortex, medulla with pyramids, renal pelvis, and renal vessels, responsible for filtration and urine production.}
\end{itemize}

\textcolor{gray}{\textbf{Test Case 3:}}
\begin{itemize}
\item \textbf{Input:} \texttt{aorta}
\item \textbf{Output:}
\textcolor{green!50!black}{Aorta, the body's largest artery originating from the left ventricle, coursing through the thorax and abdomen, featuring a thick muscular wall that delivers oxygenated blood throughout the systemic circulation.}
\end{itemize}

\textcolor{gray}{\textbf{Test Case 4:}}
\begin{itemize}
\item \textbf{Input:} \texttt{spleen}
\item \textbf{Output:}
\textcolor{green!50!black}{Spleen, a fist-sized lymphoid organ in the left upper abdomen beneath the diaphragm, featuring red and white pulp structures that filter blood, recycle iron, and support immune function.}
\end{itemize}

\vspace{0.3em}

\textcolor{teal}{\textbf{Instrument Test Cases:}}

\textcolor{gray}{\textbf{Test Case 5 (Instrument):}}
\begin{itemize}
\item \textbf{Input:} \texttt{instrument3DCOSI\_scalpels}
\item \textbf{Output:}
\textcolor{green!50!black}{Scalpels (type 1), precision cutting instruments consisting of a handle and a sharp disposable blade, used to make incisions in skin and other tissues during surgical procedures.}
\end{itemize}

\vspace{0.3em}

\subsubsection{Implementation Details}
\paragraph{Architecture}
\label{sec:architecture}
While the octree-based paradigm offers favorable scalability to higher resolutions, the dense voxelized SDF paradigm provides two properties that are particularly desirable in our setting: (i) more stable optimization on moderate-resolution volumetric data, and (ii) better compatibility with diverse conditioning signals (e.g., text prompts and partial observations) through standard latent diffusion mechanisms. We therefore adopt the dense voxelized SDF paradigm as the foundation of our baselines. Building upon these original frameworks, we further tailor them to the characteristics of medical shapes---most notably, the substantial inter-organ geometric variability and the strong emphasis on fine-grained anatomical detail. Specifically, we adopt the shape VAE from Diffusion-SDF~\cite{li2023diffusion} to encode 3D medical shapes into a compact latent space, and incorporate text prompts as conditioning signals to support multi-category medical shape generation. 
At the architectural level, we replace the standard residual blocks in the denoising UNet with text-conditioned Mixture-of-Experts (MoE)~\cite{shazeer2017outrageously} residual blocks, deployed consistently across the encoder, bottleneck, and decoder stages. Within each block, a lightweight router—implemented as a linear projection of the input text embedding—produces routing logits over four expert feed-forward networks. The top-two experts are selected via sparse soft gating, and their outputs are aggregated as a weighted sum based on normalized softmax scores, while the remaining experts are suppressed. This design enables category-aware specialization guided by the textual condition, which is particularly beneficial given the heterogeneous geometry across organ categories. To further enhance shape fidelity, we introduce a UNet adaptor module that operates on the decoded shapes and applies learned geometric refinements to recover fine-grained structural details. During training, the VAE weights are kept frozen, while the adaptor is jointly optimized with the denoising UNet in an end-to-end manner.

\paragraph{Training.}
The model is trained with a learning rate of $1 \times 10^{-3}$ for 40{,}000 steps
using DDPM with 50 diffusion steps and a batch size of 24.

\paragraph{Inference}
At inference time, we adopt DDIM sampling with 20 steps. To strengthen the guidance
effect, we introduce a \textit{recurrent guidance} mechanism: at each guided
timestep, the denoising trajectory is repeated multiple times before advancing to
the next step. Concretely, given a DDPM schedule of 50 steps, a DDIM step count
of 20, a guidance frequency of 2 (i.e., guidance is applied every 2 DDIM steps),
and a recurrence factor of 4, the model oscillates 4 times between consecutive
guided timesteps before proceeding. For example, at the first guidance interval
the trajectory alternates between timestep 49 and 46 for 4 recurrent cycles,
followed by a single standard DDIM step at 46, before moving to the next guidance
interval ($44\!\to\!41$), and so on until the final step. The resulting timestep
sequence is as follows:
\begin{equation}
\begin{split}
  &\underbrace{49, 46, \ldots, 49, 46}_{\text{guided: }49\to46},\;
   \underbrace{44, 41, \ldots, 44, 41}_{\text{guided: }44\to41},\;
   \cdots \\
  &\underbrace{9, 6, \ldots, 9, 6}_{\text{guided: }9\to6},\;
   \underbrace{4, 1, \ldots, 4, 1}_{\text{guided: }4\to1}
\end{split}
\end{equation}

\subsubsection{Time Cost Analysis}
\label{sec:time_analysis}

For the shape completion task, we evaluated the runtime under different experimental hyperparameter settings in the following environment: a single RTX 4090D GPU, CUDA 12.4, and PyTorch 2.5.1. We set the batch size to 1, ran inference on 10 cases, and report the average runtime across these 10 cases as the measure of time cost under different hyperparameter settings. Theoretically, we analyze the inference-time overhead introduced by our loss-guided optimization as follows. We adopt DDIM sampling~\cite{song2020denoising} with an interval-based guidance strategy controlled by a guidance frequency $f$, where guidance is applied at every $f$-th denoising step, yielding $S = \lfloor T / f \rfloor$ guided steps out of $T$ total DDIM steps. At each guided step, we estimate the clean prediction $\hat{x}_0$ from the current latent $x_t$ and perform $R$ recurrent gradient updates, with each update requiring one additional forward-backward pass through the denoising network, while all remaining steps follow standard DDIM denoising without any backward pass. Consequently, the total number of denoising evaluations is $T + RS$, and the number of backward passes is $RS$, giving an overall complexity of $\mathcal{O}(T + RS)$.

\subsubsection{Compared Methods}
\label{sec:search}
For each baseline method, we conduct a hyperparameter search to ensure a fair comparison, and report the best-performing results as the corresponding entries in the main table of the paper.

\definecolor{lightblue}{RGB}{220,230,255}
\begin{table}[htbp]
\centering
\caption{\textbf{Effect of CFG Method and Scale on Multiplane Shape Completion Performance.} Dice is reported in percentage (\%), while CD and UHD are scaled by $\times 100$ for better readability. All values are truncated to two decimal places.}
\label{tab:ablation_cfg_scale}
\scriptsize
\setlength{\tabcolsep}{4pt}
\begin{tabular}{@{}lccc@{}}
\toprule
Method
& Dice$\uparrow$
& CD$\downarrow$
& UHD$\downarrow$ \\
\midrule
\multicolumn{4}{c}{\textit{Scale = 3.0}} \\
\rowcolor{lightblue}
Constant     & 61.45 & 0.68 & 13.01 \\
Linear       & 49.28 & 2.53 & 21.28 \\
Exponential  & 61.37 & 0.79 & 13.24 \\
APG          & 61.22 & 0.90 & 13.40 \\
Rectified++  & 56.13 & 1.28 & 15.18 \\
Zero-Star    & 50.24 & 2.40 & 19.16 \\
\midrule
\multicolumn{4}{c}{\textit{Scale = 6.0}} \\
Constant     & 61.19 & 0.67 & 13.92 \\
Linear       & 55.33 & 1.94 & 19.02 \\
Exponential  & 63.26 & \textbf{0.55} & \textbf{12.42} \\
\rowcolor{lightblue}
APG          & \textbf{63.92} & 0.59 & 12.94 \\
Rectified++  & 56.93 & 1.04 & 14.60 \\
Zero-Star    & 54.44 & 2.15 & 17.96 \\
\midrule
\multicolumn{4}{c}{\textit{Scale = 12.0}} \\
Constant     & 36.75 & 2.09 & 28.62 \\
Linear       & 58.01 & 1.24 & 18.04 \\
Exponential  & 58.53 & 0.72 & 16.26 \\
\rowcolor{lightblue}
APG          & 60.31 & 0.71 & 17.12 \\
Rectified++  & 56.41 & 1.25 & 15.39 \\
Zero-Star    & 50.55 & 2.21 & 21.62 \\
\bottomrule
\end{tabular}
\end{table}
\definecolor{lightblue}{RGB}{220,230,255}

\begin{table}[htbp]
\centering
\caption{\textbf{Effect of CFG Method and Scale on Broken Shape Completion Performance.} Dice is reported in percentage (\%), while CD and UHD are scaled by $\times 100$ for better readability. All values are truncated to two decimal places.}
\label{tab:ablation_cfg_scale_broken}
\scriptsize
\setlength{\tabcolsep}{4pt}
\begin{tabular}{@{}lccc@{}}
\toprule
Method
& Dice$\uparrow$
& CD$\downarrow$
& UHD$\downarrow$ \\
\midrule

\multicolumn{4}{c}{\textit{Scale = 3.0}} \\
Constant     & 60.76 & 0.92 & 13.59 \\
Linear       & 50.55 & 2.33 & 20.26 \\
\rowcolor{lightblue}
Exponential  & \textbf{62.17} & \textbf{0.85} & \textbf{13.10} \\
APG          & 61.66 & 1.00 & 13.57 \\
Rectified++  & 57.93 & 1.18 & 14.60 \\
Zero-Star    & 54.50 & 1.81 & 17.38 \\
\midrule

\multicolumn{4}{c}{\textit{Scale = 6.0}} \\
Constant     & 51.22 & 1.09 & 17.95 \\
Linear       & 56.12 & 1.67 & 18.46 \\
Exponential  & 61.02 & 0.68 & 13.94 \\
\rowcolor{lightblue}
APG          & \textbf{61.56} & \textbf{0.66} & \textbf{13.59} \\
Rectified++  & 57.76 & 1.24 & 14.96 \\
Zero-Star    & 52.57 & 1.61 & 17.56 \\
\midrule

\multicolumn{4}{c}{\textit{Scale = 12.0}} \\
Constant     & 22.50 & 2.95 & 40.06 \\
Linear       & 54.48 & 1.35 & 23.86 \\
Exponential  & 46.48 & \textbf{1.17} & 25.88 \\
APG          & 45.31 & 1.24 & 26.90 \\
\rowcolor{lightblue}
Rectified++  & \textbf{57.91} & 1.22 & \textbf{14.75} \\
Zero-Star    & 48.92 & 1.38 & 21.56 \\

\bottomrule
\end{tabular}
\end{table}

\subsection{Diagnostic Applications}
\label{sec:diag_details}

\parag{Stage 1: Segmentation.} 
We adopt a 3D U-Net with feature multipliers of $[1, 2, 4, 8]$ to jointly denoise a two-channel volumetric input composed of the CT image and the mask signed distance field (SDF). The model is trained with the Adam optimizer using a learning rate of $1\times10^{-4}$ and an L1 loss over $T=300$ diffusion timesteps, with a batch size of 4, for a total of 20{,}000 steps. At test time, the model receives a degraded CT volume together with a binary observation mask, where 1 denotes observed voxels and 0 denotes missing voxels. We apply universal guidance during the last 150 denoising steps with 3 gradient iterations per guided timestep and a guidance scale of $0.5$. To improve robustness, we generate 4 independent samples with different random seeds and fuse them by averaging the SDF values before thresholding at zero.

\parag{Stage 2: Shape completion.} 
For the subsequent shape completion stage, we do not perform any retraining; instead, we directly reuse the model pretrained on MedShapeNet~\cite{li2025medshapenet} as described in Sec.~\ref{sec:exp_shape}, without any fine-tuning. The mask predicted in the previous stage is treated as an incomplete shape observation and used to guide the completion model, so that the learned shape prior can correct geometric errors such as false positives, false negatives, and incomplete structures. To make the predicted masks compatible with the shape completion model, we apply a centroid--radius transform that maps the binary mask from the KiTS23~\cite{heller2023kits21} image space to the canonical space of the shape prior: we compute the centroid and bounding radius of the foreground voxels, and then apply a global translation and isotropic scaling to align the shape to a canonical center and radius. After completion, the predicted shape is mapped back to the original space via the corresponding inverse transformation. Finally, since starting the sampling process from pure noise is often unstable for severely degraded cases with large missing regions, we adopt a DDIM inversion strategy: we select a template sample of the same anatomical category from the MedShapeNet training set, invert it to a noise state via DDIM inversion, and use the resulting noise as the initialization for guided generation, providing a more informative starting point and improving robustness under extreme degradation.

\subsection{Discussion of Generalization}
\label{sec:generalization}
In this section, we further explore the application of our method to physics simulation. By jointly generating physical properties and masks, our approach enables the prediction of physics-informed quantities, highlighting its extensibility and its ability to handle diverse tasks within a unified framework.

\subsubsection{Physics Simulation}

Recent progress in physics-informed machine learning has demonstrated the potential of deep generative models to represent and solve complex partial differential equations (PDEs). Classical numerical solvers such as Finite Element or Finite Volume Methods provide highly accurate solutions but are computationally expensive and limited by mesh resolution and domain complexity. Physics-Informed Neural Networks (PINNs)~\cite{raissi2019physics} introduced an elegant framework by embedding the PDE residuals directly into the loss function, allowing data-efficient learning while maintaining physical consistency. However, PINNs require explicit formulation of governing equations and often struggle to converge for stiff or multi-scale systems.To overcome these limitations, Neural Operators~\cite{kovachki2023neural} such as the Fourier Neural Operator (FNO)~\cite{li2021fourierneuraloperatorparametric} and Convolutional Neural Operator (CNO)~\cite{raonic2023convolutional} generalize traditional neural networks to learn mappings between function spaces, enabling resolution-invariant inference. These operator-based models have achieved significant success across a wide range of PDE problems, but they remain fundamentally deterministic and can exhibit limited expressiveness when modeling uncertainty or multi-modal physical behavior. 

In these experiments, we demonstrate how diffusion frameworks can jointly model geometry and physical quantities, providing improved flexibility compared to conventional discriminative solvers. This technique can also be used in simulation settings to iteratively steer the denoising trajectory toward physically consistent solutions using differentiable loss feedback, allowing fine-grained control over the generated outputs. By integrating ControlNet ~\cite{zhang2023adding} for structured conditioning and adapting the framework to handle both geometric and physical domains, we establish a unified generative formulation for physics simulation. 

More specifically, we jointly model the physical flow field and the geometric shape mask in a shared latent space. As discussed in Section~\ref{sec:pairwise}, for the flow field prediction task, $X$ represents the geometric shape mask and $Y$ represents the corresponding physical flow field. We conduct our evaluation on the FlowBench benchmark. To facilitate more effective latent-space encoding, we represent the FlowBench masks using signed distance function (SDF). Accordingly, $X$ more precisely refers to the SDF-encoded mask, as illustrated in the first column of Fig.~\ref{fig:vis_flow} (“SDF Mask”), while $Y$ corresponds to the physical ground truth (GT) shown in the second column.

\subsubsection{Datasets and Metrics}  
To evaluate the effectiveness of our generative formulation for physics simulation, we conduct experiments on the FlowBench benchmark~\cite{tali2024flowbenchlargescalebenchmark} for the 2D lid-driven cavity (LDC) problem. FlowBench contains flow simulation data over complex geometries, including both parametric and non-parametric shapes. It covers a broad range of flow conditions, characterized by different Reynolds and Grashof numbers, and captures diverse physical phenomena, including steady and unsteady flows, as well as forced and natural convection. Moreover, the benchmark includes both two-dimensional and three-dimensional cases. 

For evaluation, we follow the same point-wise metrics as in FlowBench~\cite{tali2024flowbenchlargescalebenchmark}, including Mean Squared Error (MSE), \(L_2\) relative error, and \(L_\infty\) relative error. These metrics are computed for all predicted physical fields, including velocity and pressure ($v_x$, $v_y$, and $p$). Specifically, let \(y\) and \(y'\) denote the predicted and ground-truth values, respectively, and let \(n\) be the total number of spatial points. The metrics are defined as
\begin{equation}
\mathrm{MSE} = \sum_{i=1}^{n}(y_i - y_i')^2,
\end{equation}
\begin{equation}
L_2 = \sqrt{\frac{\sum_{i=1}^{n}(y_i - y_i')^2}{\sum_{i=1}^{n}(y_i')^2}},
\end{equation}
\begin{equation}
L_\infty = \frac{\max_i |y_i - y_i'|}{\max_i |y_i'|}.
\end{equation}
Additionally, following FlowBench~\cite{tali2024flowbenchlargescalebenchmark}, we report the boundary-layer MSE, computed over the narrow region around the object defined by \(0 \le \mathrm{SDF} \le 0.2\). This metric better reflects the model's accuracy in capturing near-surface phenomena, which are critical for applications such as flow diagnostics, shape design, and dynamic control.
\subsubsection{Baselines}  

Similar to FlowBench~\cite{tali2024flowbenchlargescalebenchmark}, we compare our method with two established neural-operator baselines. The first is the \emph{Fourier Neural Operator} (FNO)~\cite{li2021fourierneuraloperatorparametric}, which parameterizes solution operators via global Fourier convolutions. Although highly effective for smooth parametric PDEs, FNO is inherently deterministic and can be less effective in multiscale regimes or near boundary layers. The second is the Convolutional Neural Operator (CNO)~\cite{raonic2023convolutional}, which replaces Fourier layers with localized convolutions to mitigate spectral bias and improve spatial locality. However, as a deterministic feed-forward operator, it does not provide the iterative refinement or uncertainty-aware predictions enabled by diffusion-based generative models. In addition, we compare with a strong conventional diffusion-based baseline that leverages a pretrained ControlNet~\cite{zhang2023adding}, conditioned on the geometry and flow parameters, to predict the corresponding physical quantities. Building on this, we implemented our method by following the improved guidance strategy described in Section~\ref{sec:latent}. Specifically, during the denoising process, our method additionally incorporates target-oriented SDF-encoded shape mask losses to steer the generation toward physically consistent solutions, yielding reasonable prediction results. 

\subsubsection{Implementation Details}
Because the original Stable Diffusion VAE is optimized for natural images and does not preserve fine-scale physical structure, we explore alternative encoder–decoder architectures for velocity and pressure fields and find that a deterministic autoencoder provides the most effective latent representation for diffusion-based generation.
To ensure image-space physical fidelity, we additionally apply a reconstruction loss by decoding denoised latents back to field space and supervising them against ground-truth physical quantities. Models are trained in two stages: first finetuning the chosen encoder–decoder on physics fields, then finetuning ControlNet~\cite{zhang2023adding} using both denoising and reconstruction losses. This configuration produces physically coherent predictions and consistently outperforms Neural Operator baselines.


\begin{table*}[tb]
    \centering
    \renewcommand{\arraystretch}{1}
    \setlength{\tabcolsep}{6pt}
    \scriptsize
    \caption{Quantitative comparison of different methods for physical flow field prediction}
    \label{tab:flowbench_results}

    \begin{tabular*}{\textwidth}{@{\extracolsep{\fill}}lcccccccccccc}
        \toprule
        \textbf{Model} &
        \multicolumn{3}{c}{\textbf{MSE} ($\times 10^{-4}$)$\downarrow$} &
        \multicolumn{3}{c}{\textbf{L2} ($\times 10^{-4}$)$\downarrow$} &
        \multicolumn{3}{c}{\textbf{L$\infty$} ($\times 10^{-4}$)$\downarrow$} &
        \multicolumn{3}{c}{Boundary Layer MSE ($\times 10^{-4}$)$\downarrow$} \\
        \cmidrule(lr){2-4}
        \cmidrule(lr){5-7}
        \cmidrule(lr){8-10}
        \cmidrule(lr){11-13}
        & $\mathbf{v_x}$ & $\mathbf{v_y}$ & $\mathbf{p}$
        & $\mathbf{v_x}$ & $\mathbf{v_y}$ & $\mathbf{p}$
        & $\mathbf{v_x}$ & $\mathbf{v_y}$ & $\mathbf{p}$
        & $\mathbf{v_x}$ & $\mathbf{v_y}$ & $\mathbf{p}$ \\
        \midrule

        FNO
        & 21.3 & 11.4 & 1.15e4
        & 25.5 & 27.0 & 226
        & 70.8 & 76.2 & 96.4
        & 10.8 & 7.28 & 1.03e4 \\

        CNO
        & 0.162 & 0.0941 & 179
        & 2.20 & 2.40 & 28.2
        & 16.3 & 6.90 & 89.2
        & 0.153 & 0.127 & 85.3 \\

        ControlNet~\cite{zhang2023adding}
        & 0.0400 & 0.0492 & 10.2
        & 1.10 & 1.78 & 6.72
        & 3.03 & 4.14 & 23.7
        & \textbf{0.0483} & 0.0576 & 11.0 \\
        Ours
        & \textbf{0.0223} & \textbf{0.0257} & \textbf{6.19}
        & \textbf{0.820} & \textbf{1.28} & \textbf{5.24}
        & \textbf{2.19} & \textbf{4.40} & \textbf{19.5}
        & 0.0493 & \textbf{0.0566} & \textbf{6.70} \\
        
        \bottomrule
    \end{tabular*}
\end{table*}

\begin{figure}[!t]
    \centering
    \includegraphics[width=\columnwidth]{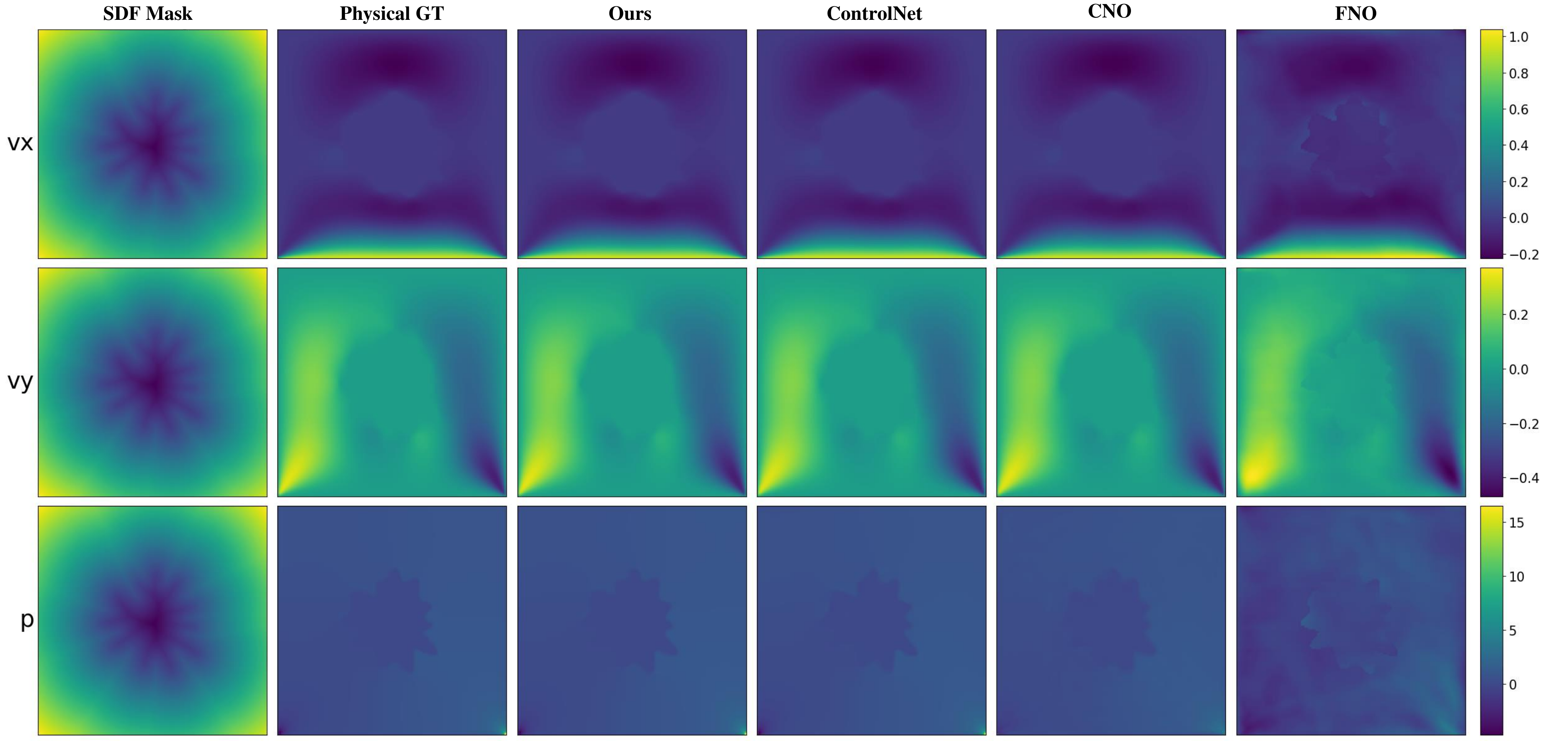}
        \caption{\textbf{Visualization of predictions across different methods for physical fields.} Each row displays the predicted $v_x$, $v_y$ (velocity components), and $p$ (pressure) fields, respectively.}
    \label{fig:vis_flow}
\end{figure}

\subsubsection{Comparative Results}
As shown in Tab.~\ref{tab:flowbench_results}, ControlNet~\cite{zhang2023adding} already provides a strong improvement over conventional operator-learning baselines, suggesting that conditional diffusion models have promising potential for physics-field prediction. Building on this, our method further improves ControlNet~\cite{zhang2023adding}  on nearly all global metrics. The gain is particularly clear for pressure reconstruction, where the MSE decreases from 10.2 to 6.19, and the corresponding L2 error decreases from 6.72 to 5.24. Similar improvements are also observed for the velocity fields, with lower MSE and L2 errors for both $v_x$ and $v_y$, as well as reduced $L_\infty$ error for $v_x$ and pressure. For boundary-layer prediction, our method remains competitive and achieves the best results on $v_y$ and $p$, while being comparable to ControlNet on $v_x$. These results indicate that our design can further enhance the predictive accuracy of diffusion-based modeling for fluid dynamics.

Even though we are not dealing with medical data anymore, this is consistent with the results obtained in medical image segmentation and shape completion. Deterministic neural operators tend to oversmooth fine-scale structures and exhibit systematic degradation on pressure fields, while diffusion-based models naturally support iterative error correction, uncertainty-aware refinement, and sharper reconstructions. By combining structured conditioning with physics-driven loss guidance, our approach yields physically coherent predictions that surpass both classical operator-based baselines and standard conditional diffusion.

\subsubsection{Visualizations}
Fig.~\ref{fig:vis_flow} shows a visual comparison of different methods for physical flow field prediction. Among the two neural-operator baselines, CNO outperforms FNO, particularly in pressure reconstruction. In comparison, diffusion-based models, including ControlNet and our method, produce visual results comparable to those of CNO. The relatively subtle differences among these three methods indicate that diffusion-based modeling is a promising approach for physics-field prediction. Together with the quantitative results in Tab.~\ref{tab:flowbench_results}, these observations highlight the strong potential of our paradigm for AI for Science tasks.

\begin{IEEEbiography}
[{\includegraphics[width=1in,height=1.6in,clip,keepaspectratio]{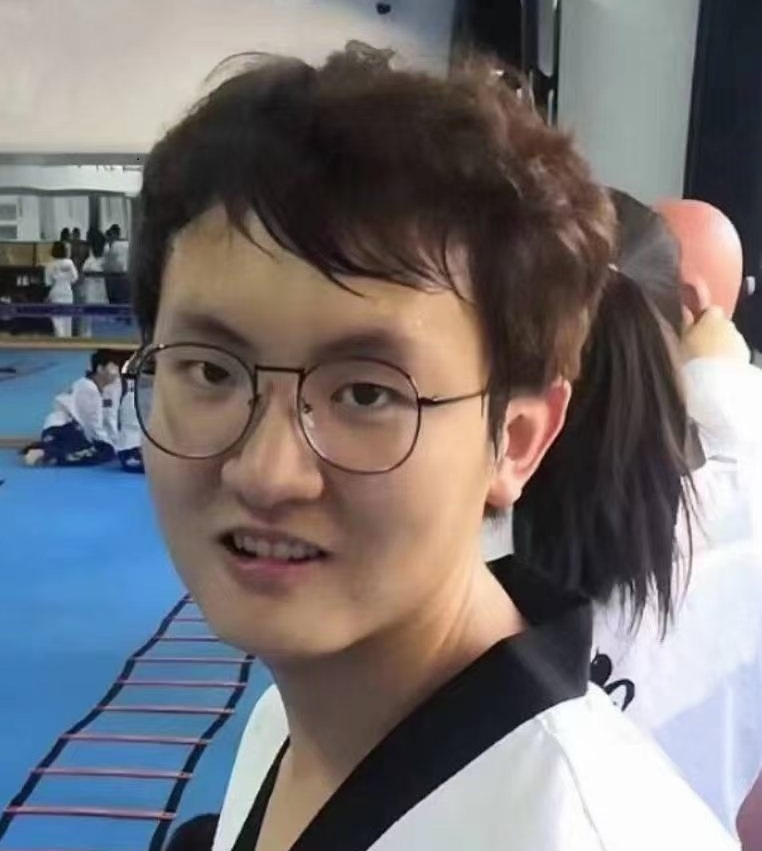}}]
{Hantao Zhang} received the B.Eng. degree in computer science and engineering from
Northeastern University, Shenyang, China, in 2022, and the M.Eng. degree in computer science and technology 
from the University of Science and Technology of China (USTC), Hefei, China, in 2025. He is currently pursuing the Ph.D. degree in the Doctoral Program in Computer and Communication Sciences (EDIC) at the École Polytechnique Fédérale de Lausanne (EPFL), Lausanne, Switzerland. His research interests include constraint-compliant controllable 
models for optimizing solutions in medical image processing, computer-aided engineering (CAE), and 3D modeling.
\end{IEEEbiography}

\begin{IEEEbiography}
[{\includegraphics[width=1in,height=1.6in,clip,keepaspectratio]{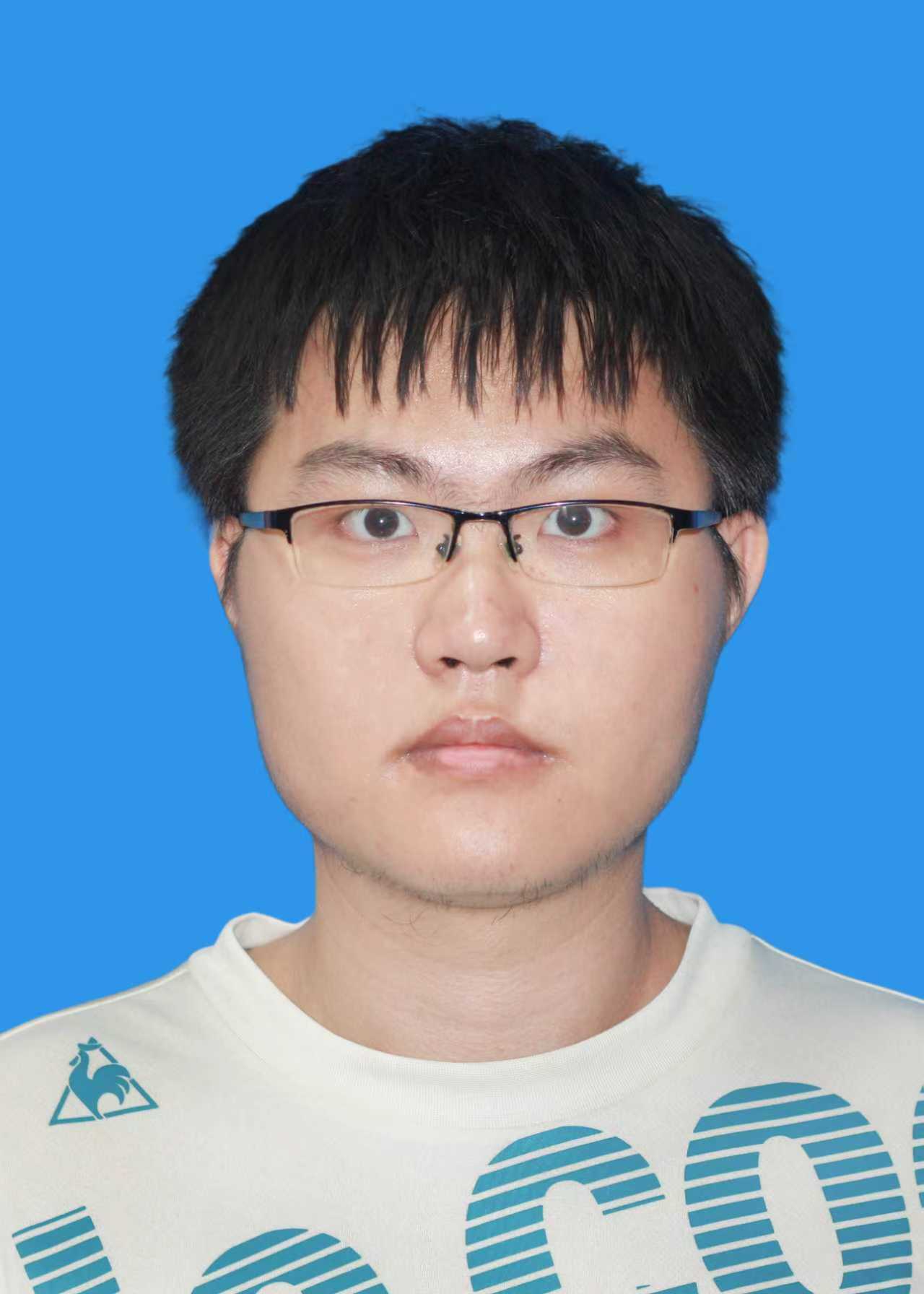}}]
{Weidong Guo} is currently working toward the PhD degree with the School of Biomedical Engineering, Fudan University, Shanghai. His research interests include 3D generative models and medical image analysis.
\end{IEEEbiography}

\begin{IEEEbiography}
[{\includegraphics[width=1in,height=1.6in,clip,keepaspectratio]{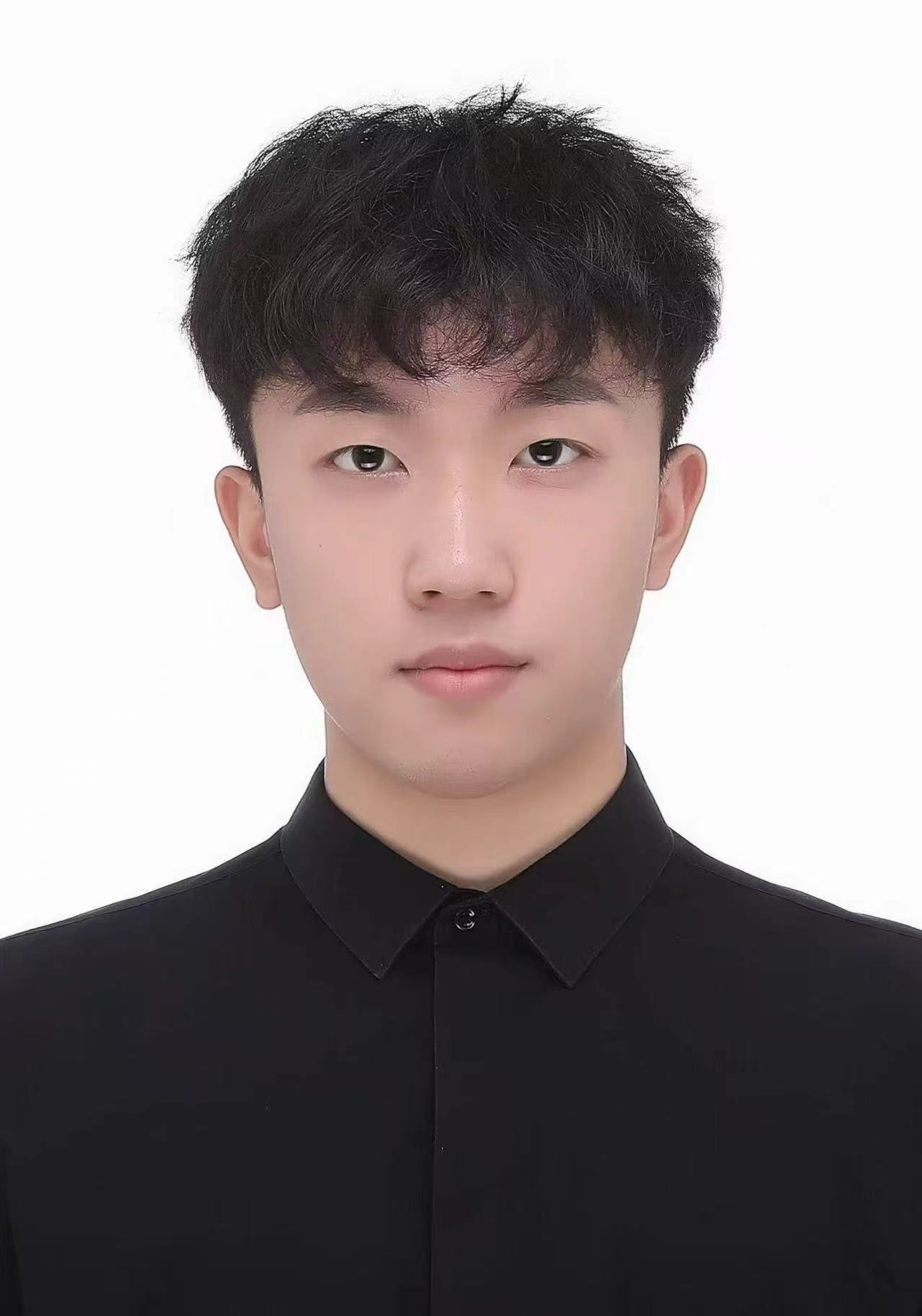}}]
{Yuhe Liu} is currently working toward the B.S. degree with the School of Artificial Intelligence, Beihang University, Beijing. His research interests focus on generative models.
\end{IEEEbiography}

\begin{IEEEbiography}
[{\includegraphics[width=1in,height=1.6in,clip,keepaspectratio]{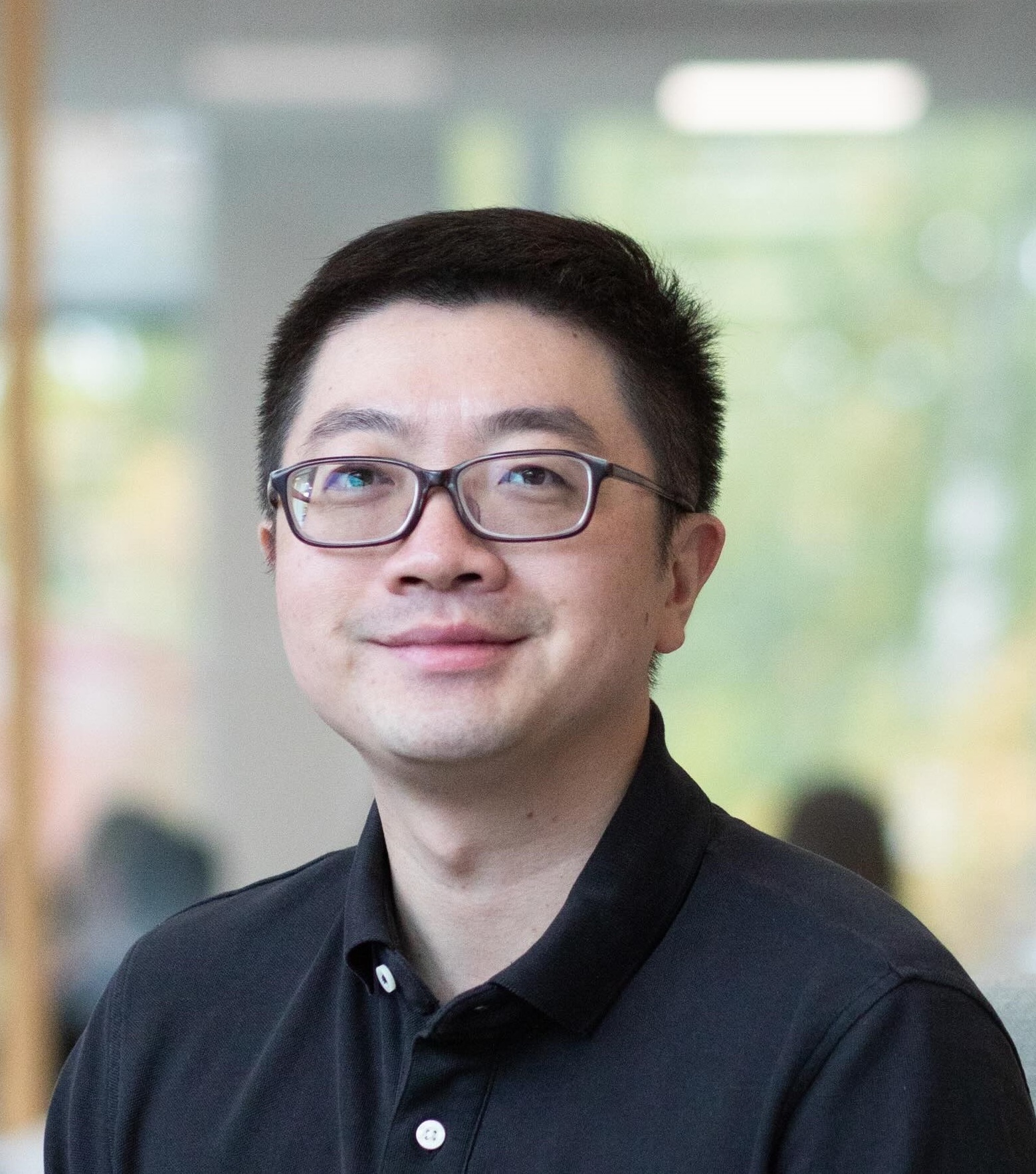}}]
{Jiancheng Yang} is a Principal Investigator at the ELLIS Institute Finland and Assistant Professor at Aalto University. He received his Bachelor's and PhD degrees from Shanghai Jiao Tong University and completed postdoctoral training at EPFL. His research focuses on AI for health, with emphasis on spatial intelligence, generative AI, and multimodal deep learning. He has authored over 60 publications and is recognized for developing MedMNIST. He is a Forbes 30 Under 30 honoree and a recipient of the WAIC Yunfan Award.
\end{IEEEbiography}

\begin{IEEEbiography}
[{\includegraphics[width=1in,height=1.6in,clip,keepaspectratio]{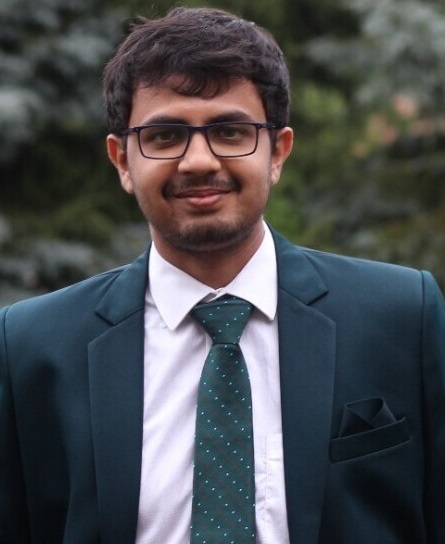}}]
{Sathvik Bhagavan} is currently pursuing a master's degree in computer science at EPFL, Switzerland. He received a bachelor's degree in mechanical engineering and computer science from the Indian Institute of Technology Kanpur in 2022. His interests include machine learning, computer vision, and data-driven approaches for scientific and engineering applications.
\end{IEEEbiography}

\begin{IEEEbiography}
[{\includegraphics[width=1in,height=1.6in,clip,keepaspectratio]{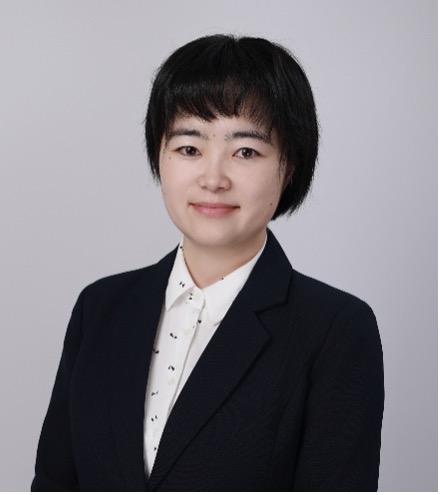}}]
{Danli Shi} received BS and MS degrees in Clinical Medicine from Shanghai Jiao Tong University and obtained the Doctor of Medicine degree from Sun Yat-sen University in 2022. She is currently a Research Assistant Professor in the School of Optometry at The Hong Kong Polytechnic University. Her research focuses on AI for healthcare.
\end{IEEEbiography}

\begin{IEEEbiography}
[{\includegraphics[width=1in,height=1.6in,clip,keepaspectratio]{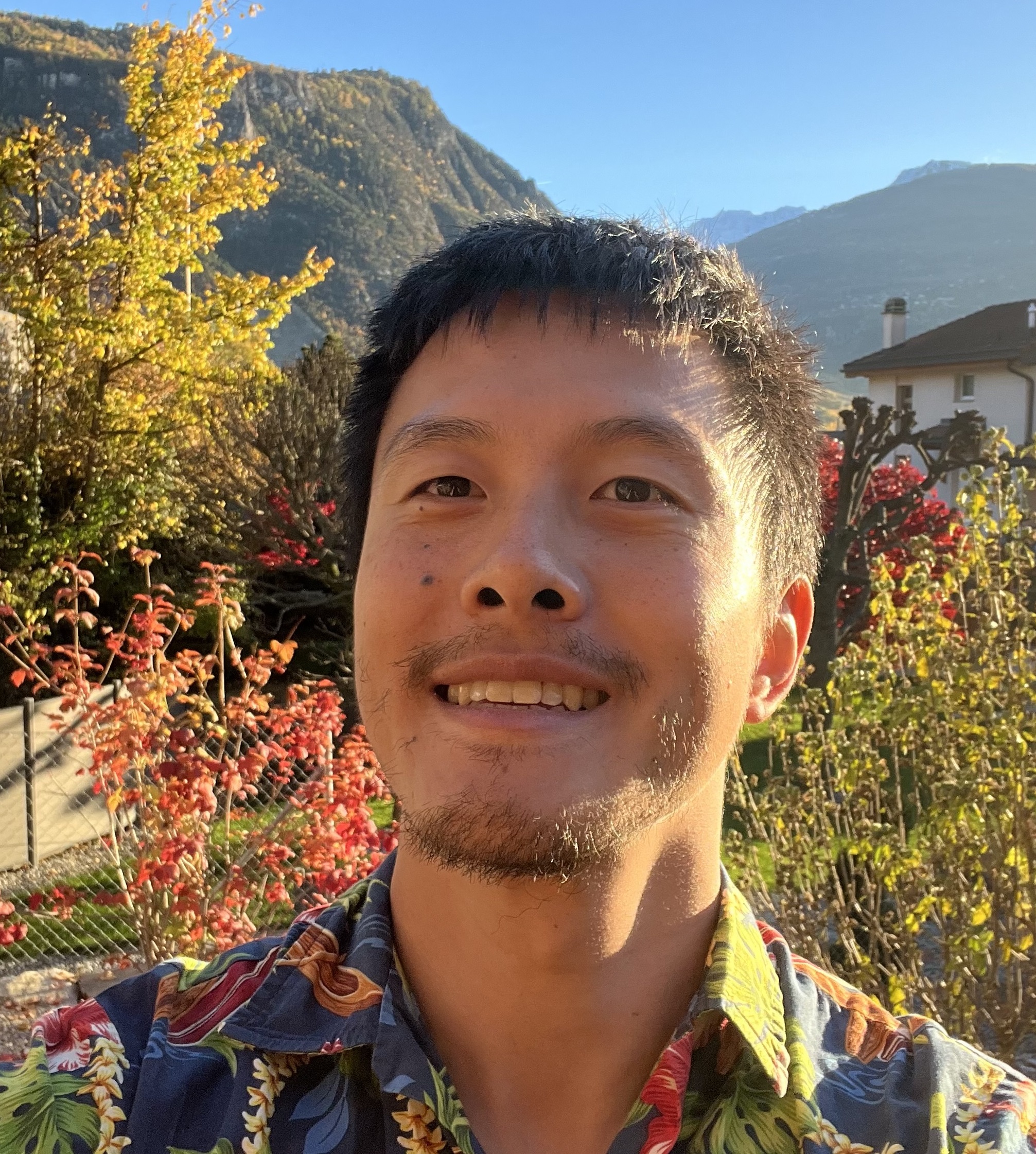}}]
{Mingda Xu} is a postdoctoral researcher at the Computer Vision Lab, École Polytechnique Fédérale de Lausanne (EPFL), Switzerland. He received the B.Comm. degree in Actuarial Studies from the University of Melbourne, the M.S. degree in Mathematics from the University of New South Wales, Sydney, and the Ph.D. degree in Robotics from the Queensland University of Technology. His research spans computer vision, machine learning, and robotics, with a current focus on optimal control and numerical optimization.
\end{IEEEbiography}

\begin{IEEEbiography}
[{\includegraphics[width=1in,height=1.25in,clip,keepaspectratio]{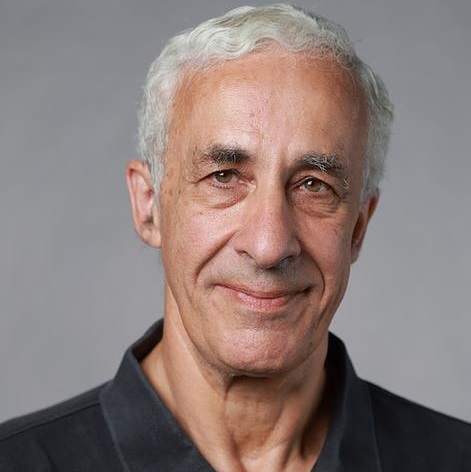}}]
{Pascal Fua} is a professor of computer science at EPFL, Switzerland, where he leads the Computer Vision Lab. He received his engineering degree in 1984 from École Polytechnique, Paris, and a PhD in computer science from the University of Orsay in 1989. His research focuses on shape and motion reconstruction from images, analysis of microscopy images, and augmented reality. He has authored or co-authored over 300 publications in refereed journals and conferences and has been awarded several ERC grants. He is an IEEE Fellow and an associate editor for IEEE Transactions on Pattern Analysis and Machine Intelligence.
\end{IEEEbiography}

\vfill

\end{document}